\documentclass[10pt,twocolumn,letterpaper]{article}

\usepackage{wacv}
\usepackage{times}
\usepackage{epsfig}
\usepackage{graphicx}
\usepackage{amsmath}
\usepackage{amssymb}

\usepackage{multirow}
\usepackage[font={small}]{caption}
\usepackage{subcaption}
\usepackage[dvipsnames]{xcolor}
\usepackage{graphicx}
\usepackage[numbers,sort]{natbib}

\wacvfinalcopy 
\usepackage[pagebackref=false,breaklinks=true,colorlinks,bookmarks=false]{hyperref}
\begin{document}

\title{RefineLoc: Iterative Refinement for Weakly-Supervised Action Localization}

\author{
Alejandro Pardo$^{1\thanks{indicates equal contribution.}}$ \quad
Humam Alwassel$^{1{\footnotemark[1]}}$ \quad
Fabian Caba Heilbron$^{2}$ \quad
Ali Thabet$^{1}$ \quad
Bernard Ghanem$^{1}$ \\
$^{1}$King Abdullah University of Science and Technology (KAUST) \quad $^{2}$Adobe Research \\
{\tt\small\{alejandro.pardo,humam.alwassel,ali.thabet,bernard.ghanem\}@kaust.edu.sa \quad caba@adobe.com} \\
{\textit{\textbf{\url{http://humamalwassel.com/publication/refineloc}}}}
}


\maketitle

\newcommand{\ahat}{\hat{\textbf{a}}}
\newcommand{\av}{\textbf{a}}
\newcommand{\bv}{\textbf{b}}
\newcommand{\cv}{\textbf{c}}
\newcommand{\dv}{\textbf{d}}
\newcommand{\uv}{\textbf{u}}
\newcommand{\vv}{\textbf{v}}
\newcommand{\x}{\textbf{x}}
\newcommand{\X}{\textbf{X}}
\newcommand{\y}{\textbf{y}}
\newcommand{\Y}{\textbf{Y}}
\newcommand{\z}{\textbf{z}}
\newcommand{\w}{\textbf{w}}
\newcommand{\W}{\textbf{W}}
\newcommand{\p}{\textbf{p}}
\newcommand{\q}{\textbf{q}}
\newcommand{\h}{\textbf{h}}
\newcommand{\A}{\textbf{A}}
\newcommand{\B}{\textbf{B}}
\newcommand{\C}{\textbf{C}}
\newcommand{\D}{\textbf{D}}
\newcommand{\F}{\textbf{F}}
\newcommand{\V}{\textbf{V}}
\newcommand{\U}{\textbf{U}}
\newcommand{\I}{\textbf{I}}
\newcommand{\PX}{\textbf{P}}
\newcommand{\mSigma}{\mathbf{\Sigma}}
\newcommand{\0}{\mathbf{0}}
\newcommand{\1}{\mathbf{1}}

\begin{abstract}
Video action detectors are usually trained using datasets with fully-supervised temporal annotations. Building such datasets is an expensive task. To alleviate this problem, recent methods have tried to leverage weak labeling, where videos are untrimmed and only a video-level label is available. In this paper, we propose RefineLoc, a novel weakly-supervised temporal action localization method. RefineLoc uses an iterative refinement approach by estimating and training on snippet-level pseudo ground truth at every iteration. We show the benefit of this iterative approach and present an extensive analysis of five different pseudo ground truth generators. We show the effectiveness of our model on two standard action datasets, ActivityNet v1.2 and THUMOS14. RefineLoc shows competitive results with the state-of-the-art in weakly-supervised temporal localization. Additionally, our iterative refinement process is able to significantly improve the performance of two state-of-the-art methods, setting a new state-of-the-art on THUMOS14.
\end{abstract}
\section{Introduction}

Weak supervision has emerged as an effective way to train computer vision models using labels that are easy and cheap to acquire. This training strategy is particularly relevant for video tasks, where data collection and annotation costs are prohibitively expensive. In this paper, our goal is to localize actions in time when no information about the start and end times of these actions is available. The lack of temporal supervision makes it challenging to train models that discriminate between action and background segments. Recent methods for weakly-supervised temporal action localization focus on learning class activation maps using soft-attention~\cite{wang_cvpr_2017}, regularizing attention with an L1 loss~\cite{nguyen_cvpr_2018}, or leveraging co-activity and multiple instance learning losses~\cite{paul_eccv_2018}. Alternatively, other methods~\cite{shou_eccv_2018, liu2019weakly} have focus on generating temporal boundaries using priors such as those encouraged by contrastive losses. All previous methods provide elegant strategies to localize actions in a weakly-supervised manner; however, they are all trained in a single shot and disregard all temporal cues. As a result, their performance lags far behind that of fully-supervised methods trained on temporal action annotations.

\begin{figure}[t]
  \centering
  \includegraphics[width=\linewidth]{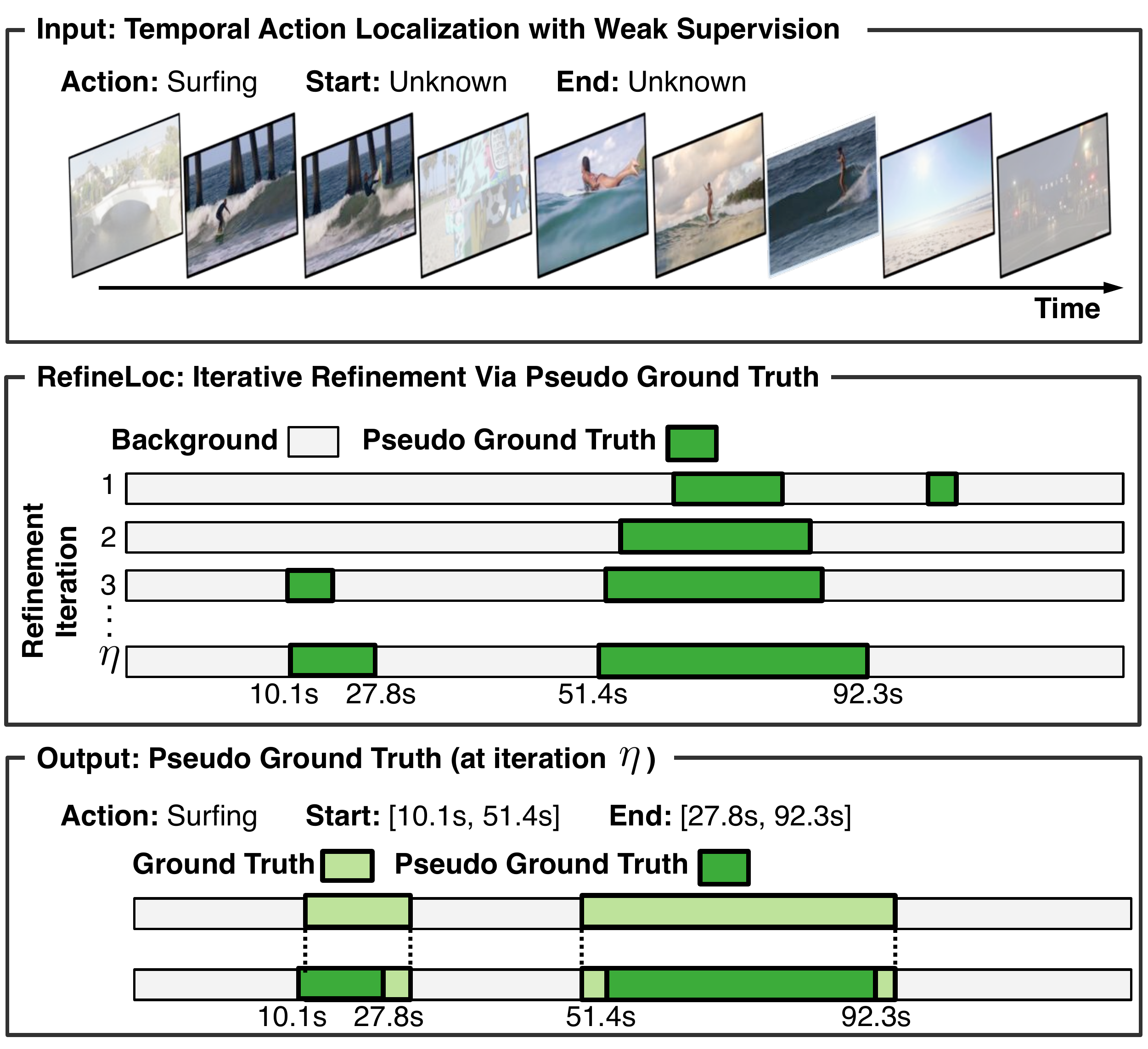}
  \caption{\textbf{Iterative Refinement for Weak Supervision.} We summarize the pseudo ground truth generation strategy used by RefineLoc. 
  \textit{Top}: The input is an untrimmed video, where only a video-level label (Surfing) is available; our goal is to correctly localize actions in time.
  \textit{Middle}: RefineLoc aims to approximate the background-foreground labels through iteratively generating pseudo ground truth (dark green boxes) using information from a weakly-supervised model. Our key idea is to use the pseudo ground truth from iteration $\eta - 1$ to supervise the detection model at iteration $\eta$.
  \textit{Bottom}: The pseudo ground truth of iteration $\eta$ (dark green) closely approximates the actual ground truth (light green).}
  \label{fig:pull_figure}
\end{figure}

In the object detection domain, refining using pseudo ground truth considerably reduces the performance gap between fully and weakly-supervised object detectors~\cite{tang_cvpr_2017, zhang_cvpr_2018}. Such pseudo ground truth refers to a set of sampled object predictions from a weakly-supervised model, which are assumed as actual object locations in the next refinement iteration. However, these methods are not directly applicable to temporal action localization. We argue this is in part due to the lack of reliable \textit{unsupervised} region proposals as in object detection.

In this paper, we propose RefineLoc, a weakly-supervised temporal localization method, which incorporates an iterative refinement strategy by leveraging pseudo ground truth. Figure \ref{fig:pull_figure} shows an example of the iterative refinement process RefineLoc employs via pseudo ground truth generation. Contrary to object detection methods, we build our refinement strategy to operate over snippet-level attention and classification modules, making it suitable for temporal localization. 

The intuition behind our iterative refinement is to leverage a weakly-supervised model, which captures decent temporal cues about actions, to annotate snippets with pseudo foreground (action) and background (no action). This pseudo ground truth is then used to train a snippet-level attention module in a supervised manner. Although such pseudo labels are noisy, it has been shown that neural networks are reasonably robust against such label perturbations~\cite{rolnick_arxiv_2017}. To avoid bias towards learning from easy examples, we randomly sample a subset of snippets for which we supervise with the pseudo labels. Our study of multiple pseudo ground truth generators shows that our simple model is competitive with the state-of-the-art. Furthermore, our iterative refinement process is generic and can be applied on top of more sophisticated models to further improve their performance.

\noindent\textbf{Contributions:} We summarize our contributions as 2-fold. \textbf{(1)} We introduce RefineLoc, an iterative refinement model for weakly-supervised temporal action localization. The model is crafted to leverage snippet-level pseudo ground truth to improve its performance over training iterations. \textbf{(2)} We show that RefineLoc's iterative refinement process improves the performance of two state-of-the-art methods, setting a new state-of-the-art on THUMOS14.\footnote{To enable reproducibility and promote future research, we have released our source code and pretrained models on our \href{http://humamalwassel.com/publication/refineloc/}{project website}.}

\section{Related Work}\label{section:related}
\noindent\textbf{Action Recognition.}
The advent of action recognition datasets such as UCF-101~\cite{dataset_ucf101}, Sports-1M~\cite{dataset_sports1m}, and Kinetics~\cite{dataset_kinetics} has fueled the development of accurate action recognition models. Traditional approaches include extracting hand-crafted representations aimed at capturing spatiotemporal features~\cite{stips, densetraj}; however, nowadays deep learning based approaches are more attractive due to their high capacity. For example, Simonyan and Zisserman~\cite{two_stream} proposed to encode spatial and temporal information with Convolutional Neural Networks. Their two-stream model represents appearance with RGB frames and motion with stacked optical flow vectors. However, the two-stream model encodes each frame independently neglecting mid-level temporal information. To overcome this drawback, Wang \etal introduced the Temporal Segment Network (TSN)~\cite{tsn}, an end-to-end framework that captures long-term temporal information. TSN along with other recent architectures (\eg I3D~\cite{i3d} and C3D~\cite{c3d}) have become the \textit{defacto} backbones for temporal action localization~\cite{piergiovanni_cvpr_2018}, action segmentation~\cite{action_segmentation}, and event captioning~\cite{event_captioning}.

\noindent\textbf{Fully-supervised Temporal Action Localization.} Multiple strategies have been developed for temporal action localization with full-supervision available at training time~\cite{alwassel_2018_actionsearch, dai_iccv_2017, gao_eccv_2018, activitynet_challenge, shou_cvpr_2017, yeung_cvpr_2016}. The first set of approaches used sliding windows combined with complex activity classifiers to detect actions~\cite{gaidon_ijcv_2013, oneata_cvpr_2014}. These methods paved the way for this type of research and established baselines and a reference for the difficulty of the problem. However, they manifested limitations regarding their run-time complexity. The second generation of methods used action proposals to speed up the search process~\cite{buch_cvpr_2017, caba_cvpr_2016, gao_iccv_2017, lin_eccv_2018, shou_cvpr_2016}. These temporal proposals aim to narrow down the number of candidate segments the action classifier examines. A third generation of approaches learn action proposals and action classifiers jointly, while back-propagating through the video representation backbone~\cite{chao_cvpr_2018, xu_iccv_2017, zhao_iccv_2017}. Finally, recent methods make use of Graph Convolutional Networks by representing videos as graphs~\cite{Zeng_2019_ICCV, xu2019gtad}.  Despite their significant performance improvements, all of these methods still rely on strong supervision that is prohibitively expensive to acquire.

\noindent\textbf{Weakly-supervised Temporal Action Localization.} The challenge in this task is to learn to discriminate between background and action segments without having explicit temporal training samples, but instead, only a coarse video-level label. The first methods proposed solutions consisting of hiding video regions to encourage their model to discover discriminative parts~\cite{kumar2017hide}, and a soft-attention layer to focus on snippets that boost the video classification performance~\cite{wang_cvpr_2017}. Similarly,~\cite{nguyen_cvpr_2018} proposed an attention layer regularized with an L1 loss. Other works explored different alternatives such as co-activity loss combined with a multiple instance learning loss~\cite{paul_eccv_2018} and action proposal generation using contrast cues among action classification predictions~\cite{shou_eccv_2018, liu2019weakly}. With the end goal of addressing the lack of temporal information, other works have innovated strategies such as incorporating temporal structure~\cite{yu2019temporal}, modeling background~\cite{nguyen2019weakly, basnet_aaai20}, using extra supervision (\eg action count~\cite{3cnet}), or single-frame label~\cite{ma2020sfnet}. More recent methods have tried to reduce the supervision level by using self-supervised techniques~\cite{Jain_2020_CVPR}. Our work builds upon these ideas and complements them with a key insight: leveraging pseudo labels while \emph{iteratively} training the model. 

\noindent\textbf{Weak Supervision and Pseudo-labeling in Vision Tasks.} 
Weak supervision has been widely studied in other vision tasks such as object detection~\cite{bilen_cvpr_2016, oquab_cvpr_2015, shi_iccv_2017, singh_iccv_2017}, semantic segmentation~\cite{papandreou_iccv_2015, xu_cvpr_2015}, or other video tasks~\cite{escorcia_arxiv_2018, huang_cvpr_2018, huang_eccv_2016, richard_cvpr_2017}. For video tasks, a variety of weak supervision cues have been used including movie scripts~\cite{laptev2008learning, duchenne_iccv_2009, kuehne2017weakly, miech_iccv_2017}, action ordering priors ~\cite{bojanowski_eccv_2014, richard2018neuralnetwork, ding2018weakly, chang_cvpr_2019}, and different levels of supervision~\cite{cheron2018flexible}. These video related solutions have proposed innovative ways to reduce labeling expense; however, they still require laborious annotations (\eg action spots) or privileged information (\eg transcripts) that is difficult to obtain beyond a controlled setting. Concerning pseudo-labeling, it has been used to design state-of-the-art methods for weakly-supervised object detection~\cite{tang_cvpr_2017, zhang_cvpr_2018}, train image classification backbones~\cite{caron2018deepcluster, caron2019self-pseudo}, and build pose detectors~\cite{neverova2019slim}. These works have inspired our model, which addresses challenges unique to the weakly supervised temporal action localization task, namely the presence of only a sparse supervision signal (video-level action category) and of highly similar context surrounding the action~\cite{alwassel_eccv_2018}.
\begin{figure*}[t]
  \centering
  \includegraphics[width=\linewidth]{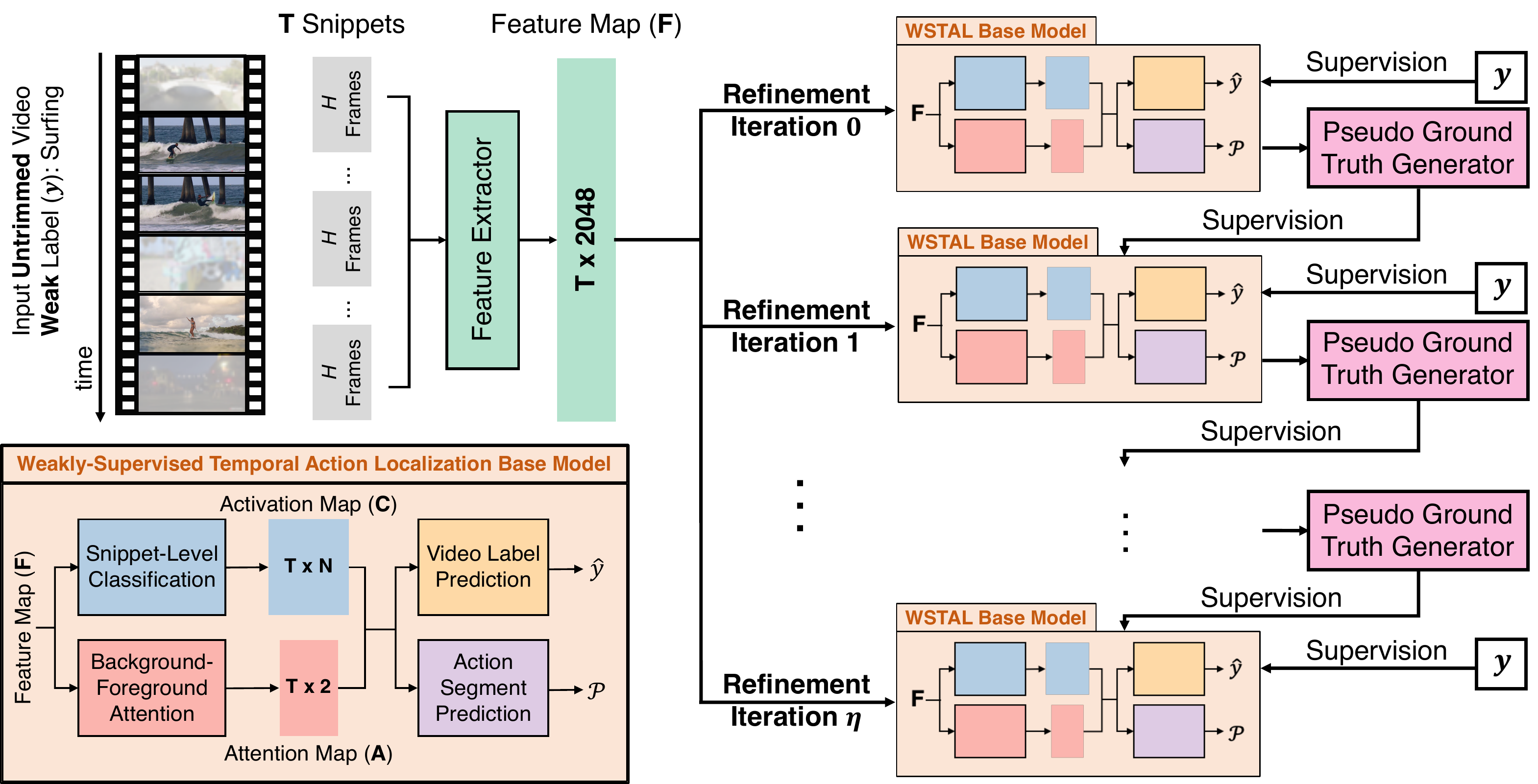}
  \caption{\textbf{Overview of RefineLoc Architecture.} Given an untrimmed video with only a weak label $\y$, we extract spatio-temporal feature map $\F$ from $T$ non-overlapping $\textit{H}$-frame-long snippets (\textit{top left}). We feed $\F$ to an iterative refinement process (\textit{right}). At each iteration, $\F$ passes through our WSTAL base model (\textit{bottom left}) to compute a snippet-level activation map ($\C$) and a background-foreground attention map ($\A$). Both $\A$ and $\C$ are used to predict the video label $\hat{\y}$ and later to produce action segment predictions $\mathcal{P}$. At iteration $0$, the pipeline is supervised using only $\y$. Subsequent iterations use both $\y$ and pseudo ground truth generated from the previous iteration. We stop the refinement process after $\eta$ iterations.
  }
  \label{fig:method}
\end{figure*}

\section{RefineLoc}
\label{section:model}

In this section, we discuss our RefineLoc architecture, the pseudo ground truth label generation, and the iterative refinement process. The input to our model is an untrimmed video and the expected output is a set of action segment predictions. RefineLoc is supervised on weak labels (\ie video-level action labels) and \textbf{does not use} any temporal annotations of action instances. RefineLoc has two main components: a weakly-supervised temporal action localization (WSTAL) base model (Subsection \ref{subsection:wstal}) and an iterative refinement process (Subsection \ref{subsection:iterative_refinement}). Based on a trained WSTAL model, we generate pseudo background-foreground ground truth labels. We use these pseudo labels to supervise the training of a new WSTAL model. We repeat the process for $\eta$ iterations to progressively improve the pseudo ground truth and refine the final action prediction segments. Figure \ref{fig:method} illustrates our approach.

\subsection{WSTAL Base Model} \label{subsection:wstal}
The input to WSTAL is an untrimmed video, while the output is temporal action segment predictions. First, WSTAL extracts features form $T$ non-overlapping snippets, which are then fed into both a snippet-level action classifier and a background-foreground attention module. Then, WSTAL combines the class activation and attention maps to produce a video label prediction $\hat{\y}$. 
During training, we supervise WSTAL with a cross-entropy loss between the ground truth video label $\y$ and the predicted label $\hat{\y}$. Finally, we post-process the learned class activation and attention maps to produce action segment predictions. In what follows, we discuss the details of each module in WSTAL.

\noindent\textbf{Feature Extraction Module.}
To compare with other works, we use two feature extractor backbones: TSN~\cite{tsn} (pretrained by UntrimmedNets~\cite{wang_cvpr_2017}) and I3D~\cite{i3d} (pretrained on Kinetics~\cite{dataset_kinetics}).
We split the input untrimmed video into $T$ non-overlapping $H$-frame-long clip snippets ($15$ for TSN and $16$ for I3D). We transform each snippet into a $2048$-dimensional feature vector by concatenating the two $1024$-dimensional activation vectors from the global pooling layer of each stream. Thus, this module outputs a $T \times 2048$ feature map $\F$.

\noindent\textbf{Snippet-Level Classification Module.}
This module receives the feature map $\F$ and produces a $T \times N$ class activation map $\C$, where $N$ is the number of action classes ($100$ classes in ActivityNet v1.2~\cite{dataset_activitynet} and $20$ in THUMOS14~\cite{dataset_thumos14}). It consists of a multi-layer perceptron (MLP) with $L$ Fully-Connected (FC) layers interleaved with ReLU activation functions. We reduce the size of each hidden layer by $2$, which makes the last layer of size $\frac{2048}{2^{L-1}} \times N$. 

\noindent\textbf{Background-Foreground Attention Module.} The objective of this module is to learn attention weights for each snippet to suppress the background snippets and to focus on foreground snippets. It transforms $\F$ into a $T \times 2$ background-foreground attention map $\A$. Similar to the Snippet-Level Classification Module, it consists of an MLP with $L$ FC layers interleaved with ReLUs. Each hidden layer size is reduced by half, making the last FC layer of size $\frac{2048}{2^{L-1}} \times 2$. Other weakly-supervised action localization methods~\cite{nguyen_cvpr_2018, nguyen2019weakly, liu2019weakly, shou_eccv_2018, wang_cvpr_2017, 3cnet} employ attention modules in their models. While we share a similar motivation, our attention module is different from theirs in one key aspect: their attention modules are only supervised by the video-level label for the purpose of improving the video classification, while our attention is supervised by both the video-level label and a set of pseudo background-foreground labels with the goal of improving the action segment localization. Subsection \ref{subsection:iterative_refinement} details the pseudo ground truth label generation process. Unlike previous methods with a scalar attention, we model the attention explicitly with two values, one for foreground and one for background. We chose to do so because our method uses supervision directly on the attention values. Thus, instead of learning the attention with a logistic-regression loss, we learn it as a binary classification problem. We compare learning a scalar attention via logistic regression against our proposed two dimensional attention in the \textit{supplementary material}. 

\noindent\textbf{Video Label Prediction Module.}
This module combines $\C$ and $\A$ to generate an $N$-~dimensional probability vector $\hat{\y}$ for the video label. Specifically, we pass $\C$ through a softmax layer across the class dimension to get $\bar{\C}$ and pass $\A$ through two softmax layers. The first softmax layer operates across the background-foreground dimension to produce $\bar{\A}^{bf}$, 
while the second softmax layer operates across the time dimension (across snippets) of the foreground attentions in $\bar{\A}^{bf}$ to produce $\bar{\A}^{time}$ as follows:
\begin{align}
    &\bar{\A}_{t,i}^{bf}   = \frac{\exp(\A_{t,i})}{\exp(\A_{t,1}) + \exp(\A_{t,2})} \label{equation:Attentionbf},\\ 
    &\bar{\A}_{t,i}^{time} = \frac{\exp(\bar{\A}_{t,i}^{bf})}{\sum_{t'=1}^{T} \exp(\bar{\A}_{t',i}^{bf})} \label{equation:Attentiontime}.
\end{align}

\noindent Here, we use $\bar{\A}^{bf}$ as the network's predictions for the snippet-level background-foreground pseudo ground truth supervision (Subsection \ref{subsection:iterative_refinement}). Note that in Equations \ref{equation:Attentionbf} and \ref{equation:Attentiontime}, $i=1$ refers to background while $i=2$ refers to foreground. 
Finally, this module computes the video-label prediction as 
$\hat{\y} = \sum_{t=1}^{T} (\bar{\A}^{time}_{t} \cdot \bar{\C}_{t})$,
where $\bar{\A}_{t}^{time}$ and $\bar{\C}_{t}$ are the foreground attention value and class activation vector of the $t^{\text{th}}$ snippet. 
The video-label prediction uses a soft attention mechanism to emphasize the class activations of snippets with higher attention values.

\noindent\textbf{Action Segment Prediction Module.}
This module post-processes $\bar{\A}^{bf}$ and $\bar{\C}$ to produce a set of action segment predictions $\mathcal{P}$. 
First, we filter out snippets for which the background attention value is greater than a threshold $\alpha_A$. Then, we consider only the top-$k$ classes in $\hat{\y}$. For each top class $n$, we filter out snippets that have classification score lower than a threshold $\alpha_C$. Then, we generate contiguous segments by grouping snippets that are separated by at most one filtered-out (background) snippet. 
We do so to overcome noise in the filtering process and connect segments that are close to each other. This process can be done in other and more sophisticated ways, however we keep the simplicity of the base model and rely mainly on our iterative process. We assign to each predicted segment ($t_1$, $t_2$) the label $n$ and the score $s$,
\begin{align}
    s = \frac{1}{t_2 - t_1 + 1} \sum_{t=t_1}^{t_2} \left(\bar{\A}^{time}_{t} + \bar{\C}_{t,n}\right) + \hat{\y}_n.
\end{align}

\noindent where $\hat{\y}_n$ is the video-level predictions score for the $n^{\text{th}}$ class. Note that each prediction that comes from the $n^{\text{th}}$ top-$k$ labels, has a different score $s$. Finally, to encode temporal context and deal with the ambiguity of action boundaries \cite{alwassel_eccv_2018, sigurdsson_iccv_2017}, we inflate segments by $2$ snippets at both ends.

\subsection{Iterative Refinement Process}
\label{subsection:iterative_refinement}
Let $\mathcal{M}_0$ be the WSTAL base model trained using the weak video labels only. We iteratively refine this base model and its action predictions by introducing supervision on the background-foreground attention module using snippet-level pseudo ground truth labels. Let $\mathcal{G}^{\mathcal{M}_\eta}$ be the pseudo ground truth generation function that uses information from $\mathcal{M}_\eta$ (the trained WSTAL base model after iteration $\eta$) to map each snippet to a pseudo background-foreground label. At iteration $\eta+1$, we train a new WSTAL base model $\mathcal{M}_{\eta+1}$ on the joint loss of the video-level label and the snippet-level pseudo ground truth labels from $\mathcal{G}^{\mathcal{M}_\eta}$. 
Specifically, we compute the loss for $\mathcal{M}_{\eta + 1}$ on a given video in the following way,
\begin{align} \label{equation:loss}
    \text{loss} = \mathcal{L}\left(\hat{\y}, \y \right) + \beta \frac{1}{T} \sum_{t=1}^{T}  \mathcal{L}\left(\bar{\A}^{bf}_{t}, \mathcal{G}^{\mathcal{M}_\eta}(t)\right),
\end{align}
where $\mathcal{L}$ is the cross-entropy loss and $\beta$ is a trade-off coefficient to balance the loss signal of the pseudo ground truth with that of the video label. Note that the second cross-entropy loss is class-weighted to alleviate the imbalance in background and foreground pseudo labels.

\noindent\textbf{Pseudo Ground Truth Generation.} Intuitively, to obtain the maximum gain from the iterative refinement process, we want a pseudo ground truth generator that provides the closest approximation to the \emph{true} snippet-level background-foreground ground truth labels, \ie it should minimize the mislabelling rate. In order to overcome any possible bias learned by the pseudo ground truth generator and inspired by~\cite{kumar2017hide}, we only fixate on a portion of the pseudo ground truth in a process we call \textit{pseudo ground truth sampling}: at the start of each refinement iteration, we randomly sample a percentage $S$ of snippets for which we apply the pseudo ground truth loss. We consider five different pseudo ground truth generation strategies and study their effects on the localization performance (Subsection \ref{subsection:ablation}).

\noindent\textbf{(1)} \textit{\textbf{Uniformly Random Generator:}} This generator assigns a uniformly random pseudo label to each snippet.

\noindent\textbf{(2)} \textit{\textbf{Distribution Aware Generator:}} This generator gives, with a biased probability, a random pseudo ground truth label to each snippet. The biased probability is equal to the average ratio of actual foreground to background snippets. This generator relies on information (namely the ratio) that requires access to strong temporal annotations. Thus, it does not align with the weakly-supervised setting, but we include it as a baseline reference only.

\noindent\textbf{(3)} \textit{\textbf{Class Activation-Based Generator:}} This generator selects the pseudo ground truth label for a snippet $t$ by thresholding its maximum class score, $\max(\bar{\C}_{t})$.

\noindent\textbf{(4)} \textit{\textbf{Attention-Based Generator:}} This generator produces the pseudo ground truth label for a snippet $t$ by thresholding its foreground attention value, $\bar{\A}^{time}_{t}$.

\noindent\textbf{(5)} \textit{\textbf{Segment Prediction-Based Generator:}} This generator assigns pseudo labels based on the set of prediction segments $\mathcal{P}$. A snippet is given a pseudo foreground label if it is covered by a segment prediction and a pseudo background label otherwise. We use this generator in our final model due to its attractive performance gain.

\section{Experiments}\label{section:experiments}
\subsection{Datasets and Evaluation Metric}
We conduct our experiments on ActivityNet v1.2~\cite{dataset_activitynet} and THUMOS14~\cite{dataset_thumos14}. Both datasets consist of untrimmed videos with (weak) video-level action labels and have (strong) temporal annotations of action instances. However, \textit{we discard the strong annotations during training}.

\noindent\textbf{THUMOS14}~\cite{dataset_thumos14}. This dataset has $1010$ validation and $1574$ testing videos annotated with $101$ sport-related action classes at the video-level. Among these videos, only $200$ validation and $213$ testing videos have temporal annotations for $20$ sport actions. As in prior work~\cite{gao_eccv_2018, zhao_iccv_2017}, we only consider these $20$ classes, use the $200$ validation videos to train, and use the $213$ testing videos to evaluate performance.

\noindent\textbf{ActivityNet v1.2}~\cite{dataset_activitynet}. This dataset has $9682$ untrimmed videos annotated with $100$ activity classes. It is split into training, validation, and testing subsets, where the testing subset labels are withheld for an annual challenge. Following other methods~\cite{paul_eccv_2018, shou_eccv_2018}, we use the training subset ($4819$ videos) to train and the validation subset ($2383$ videos) to test the performance. ActivityNet is a challenging dataset due to its large-scale nature and, unlike THUMOS14, its diverse classes ranging from household activities to sports.

\noindent\textbf{Evaluation Metric.} We compare methods according to mean Average Precision (mAP). We report mAP at multiple temporal Intersection-over-Union (tIoU) thresholds. We take the average mAP across tIoU thresholds $0.5$:$0.05$:$0.95$ as the main metric for ActivityNet v1.2 and the mAP at tIoU threshold $0.5$ as the evaluation metric for THUMOS14.

\subsection{Implementation Details}
We extract features from two different architectures: an I3D model \cite{i3d}, and the same pre-trained TSN \cite{tsn} model used in AutoLoc \cite{shou_eccv_2018}, with 16 and 15 number of frames per snippet ($H$), respectively. We choose $L=2$ layers for the snippet-level classification and background-foreground attention modules. In the action segment prediction module, we set $(\alpha_A,\alpha_C)$ to $(0.5, 0.005)$ for ActivityNet and $(0.5, 0.35)$ for THUMOS14. We consider the top-$2$ labels when generating segment predictions in both datasets. At every iteration, we randomly sample $S=80\%$ of the pseudo labels. Finally, we use an initial learning rate of $10^{-4}$ for ActivityNet and $10^{-3}$ for THUMOS14, and decay the learning rate by $0.9$ when the validation loss saturates. We train for $50$ epochs per refinement iteration and pick the best model with the lowest validation loss from Equation~\ref{equation:loss}.

\subsection{Ablation Study}
\label{subsection:ablation}

In this subsection, we present multiple ablation studies motivating the design choices for our RefineLoc approach. First, we study the performance of several pseudo ground truth generators and the influence of the loss trade-off coefficient $\beta$ (Equation \ref{equation:loss}) on the performance of each generator. Afterwards, we analyze how our model's performance changes from one refinement iteration to the next. Finally, we present a diagnosis study (using the DETAD~\cite{alwassel_eccv_2018} diagnostic tool) of the detection results before and after our iterative refinement process. We present all the studies in this subsection using ActivityNet v1.2~\cite{dataset_activitynet} dataset along with I3D features. For all the experiments in this section we report average mAP at tIoU thresholds $0.5$:$0.05$:$0.95$. Refer to the \textit{supplementary material} for the study results on ActivityNet v1.2 using TSN features as well as on THUMOS14~\cite{dataset_thumos14} using I3D and TSN features.

\noindent\textbf{Effects of the Pseudo Ground Truth Generator and the Loss Trade-off Coefficient $\beta$.}
Table \ref{table:ablaiton_study_anet_generator_and_beta} summarizes the best average mAP performance using the five generators with five $\beta$ values. The baseline model $\mathcal{M}_{0}$ ($\beta = 0$) achieves $9.66\%$ average mAP at tIoU=$0.5$:$0.05$:$0.95$. We observe a performance improvement over $\mathcal{M}_0$ across all generator types and $\beta$ values. This shows the effectiveness of our iterative refinement process. Moreover, we observe that the segment prediction-based generator is the best among the five generators. We hypothesize that this generator is better, since it has access to information from both the class activation and attention maps. Moreover, $\beta=4$ strikes the best balance between the video label loss and the background-foreground pseudo ground truth loss. We observe similar results on THUMOS14: the best generator is the segment prediction-based one and the best $\beta$ is $4$. 

\begin{table}[t]
    \small
	\centering
    \tabcolsep=0.15cm
	\begin{tabular}{ l | c c c c c c}
		\hline
		
		\textbf{Pseudo Ground}    & \multicolumn{6}{c}{\textbf{$\beta$}}\\
		\textbf{Truth Generator}  & 0    & 1    & 2    & 4    & 8    & 16   \\
		
		\hline
            				
        Uniform Random & \multirow{5}{0.9em}{\rotatebox{90}{---\ \ 9.66\ \ ---}} & 9.66 & 9.66 & 9.66 &  9.66 & 9.66 \\
        Distribution Aware  &  & 17.39 & 19.10 & 20.00 & 17.73 & 18.30 \\
        Class Activation    &  & 23.09 & 23.02 & 22.93 & 22.86 & 22.85 \\
        Attention           &  & \textbf{23.15} &  23.13 & 22.97 & 23.00 & 22.94 \\
        Segment Prediction  &  & 23.04 & \textbf{23.15} & \textbf{\underline{23.24}} & \textbf{23.11} & \textbf{23.09} \\
		\hline
	\end{tabular}
	\caption{\textbf{Effects of pseudo ground truth generator and loss trade-off coefficient $\beta$ on ActivityNet v1.2}.
	The segment prediction-based generator with $\beta=4$ shows the highest performance (underlined). Bold numbers mark the best performing generator for each $\beta$.
	}
	\label{table:ablaiton_study_anet_generator_and_beta}
\end{table}

\begin{table}[t!]
    \small
	\centering
    \tabcolsep=0.1cm
	
	\begin{tabular}{ l | c c c c c c}
		\hline
		
		Refinement Iteration    & 0 & 1 & 2 & 3  & 4 & 5   \\
		
		\hline
            				
        \textbf{RefineLoc}  & 9.66  & 19.14 & 22.66 & \textbf{23.24} & 22.94 & 22.95 \\
		\hline
	\end{tabular}
	\caption{\textbf{Effects of refinement}. We show the gain from our iterative refinement on ActivityNet v1.2. Note the significant improvement over iterations: $13.58\%$ in $3$ iterations. 
	}
	\label{table:ablaiton_study_anet_iterations}
\end{table}

\noindent\textbf{Performance over Refinement Iterations.}
Table \ref{table:ablaiton_study_anet_iterations} shows the evolution of RefineLoc's performance across five refinement iterations. We obtain the highest performance (average mAP of $23.24\%$) after $\eta = 3$ iterations. This is a significant $13.64\%$ increase over our baseline model $\mathcal{M}_0$ (iteration $0$ in the table). We also see that refining $\mathcal{M}_0$ for a single iteration boosts the performance by $9.48\%$. This clearly shows the effectiveness of leveraging the pseudo ground truth labels during training. We observe similar results on THUMOS14: the best performance is achieved after $\eta=3$ refinement iterations.

\begin{table}[t]
    \small
	\centering
	\tabcolsep=0.27cm
	\begin{subtable}{0.9\linewidth}
	    \centering
	    \caption{Methods using TSN features} \label{table:activitynet_sota_tsn}
	    \vspace{-6pt}
    	\begin{tabular}{l | c c c | c }
    		\hline
    		{Method} & {0.5} & {0.75} & {0.95} & {Avg.}\\
    		\hline 
    		{UntrimmedNets~\cite{wang_cvpr_2017}} & {7.4} & {3.2} & {0.7} & {3.6}\\
    		{AutoLoc~\cite{shou_eccv_2018}} & {27.3} & {15.1} & {3.3} & {16.0}\\
    		{TSM~\cite{yu2019temporal}} & {28.3} & {17.0} & {3.5} & {17.1} \\
            {CMCS~\cite{liu2019completeness}} & {33.9} & {19.9} & {5.1} & {20.5} \\
            {CleanNet~\cite{liu2019weakly}} & {37.1} & {20.3} & {5.0} & {21.6} \\
    		\hline
            {\textbf{RefineLoc} ($\eta = 0$)} & {25.8} & {11.5} & {2.8} & {13.3} \\
            {\textbf{RefineLoc} ($\eta = 5$)} & \textbf{{38.8}} & \textbf{22.2} & \textbf{5.3} & \textbf{{23.2}} \\
    		\hline
    	\end{tabular}
	\end{subtable}
	\\
	\tabcolsep=0.3cm
	\begin{subtable}{0.9\linewidth}
	    \centering
	    \vspace{3pt}
    	\caption{Methods using I3D features} \label{table:activitynet_sota_i3d}
     	\vspace{-6pt}
    	\begin{tabular}{l | c c c | c }
    		\hline
    		{Method} & {0.5} & {0.75} & {0.95} & {Avg.}\\
    		\hline 
            {W-TALC~\cite{paul_eccv_2018}} & {37.0} & {-} & {-} & {18.0}\\
            {3C-Net \cite{3cnet}} & {35.4} & {-} & {-} & {21.1}\\ 
            {3C-Net$\dagger$ \cite{3cnet}} & {37.2} & {-} & {-} & {21.7}\\
            {CMCS~\cite{liu2019completeness}} & {36.8} & {22.0} & {5.6} & {22.4} \\
            {BaS-Net~\cite{basnet_aaai20}} & {38.5} & \textbf{24.2} & \textbf{5.6} & \textbf{24.3} \\
            \hline
            {\textbf{RefineLoc} ($\eta = 0$)} & {19.2} & {8.0} & {2.3} & {9.7} \\
            {\textbf{RefineLoc} ($\eta = 3$)} & \textbf{38.7} & {{22.6}} & {{5.5}} & {{23.2}} \\
    		\hline
    	\end{tabular}
	\end{subtable}
	\vspace{6pt}
	\caption{\textbf{State-of-the-art Weak Supervision on ActivityNet v1.2}. RefineLoc outperforms other methods using TSN features (a) and is competitive using I3D features (b). 3C-Net$\dagger$~\cite{3cnet} uses number of instances per video as extra supervision.
	}
	\label{table:activitynet_sota}
\end{table}

\begin{table}[t]
    \small
	\centering
	\tabcolsep=0.2cm
	\begin{subtable}{\linewidth}
	    \centering
	    \caption{Methods using TSN features} \label{table:thumos14_sota_tsn}
	    \vspace{-6pt}
	    \begin{tabular}{l | c c c  c c}
    		\hline
    		{Method} & {0.3} & {0.4} & {0.5} & {0.6} & {0.7}\\
    		\hline 
    		{UntrimmedNets~\cite{wang_cvpr_2017}} & {28.2} & {21.1} & {13.7} & {-} & {-} \\
    		{W-TALC~\cite{paul_eccv_2018}} & {32.0} & {26.0} & {18.8} & {-} & {6.2} \\
            {CMCS~\cite{liu2019completeness}} & {37.5} & {29.1} & {19.9} & {12.3} & {6.0} \\ 
    		{AutoLoc~\cite{shou_eccv_2018}} & {35.8} & {29.0} & {19.9} & {12.3} & {6.0} \\
    		{CleanNet~\cite{liu2019weakly}} & {37.5} & {29.1} & {23.9} & {13.9} & {7.1} \\ 
    		{BaS-Net~\cite{basnet_aaai20}} & \textbf{42.8} & \textbf{34.7} & \textbf{25.1} & \textbf{17.1} & \textbf{9.3} \\
    		\hline
    		{\textbf{RefineLoc} ($\eta = 0$)} & {7.0} & {4.2} & {2.9} & {1.3} & {0.6} \\
    		{\textbf{RefineLoc} ($\eta = 4$)} & {36.1} & {29.6} & {22.6} & {12.1} & {5.8} \\
            \hline
        \end{tabular}
    \end{subtable}
	\begin{subtable}{\linewidth}
	    \centering
	    \vspace{3pt}
	    \caption{Methods using I3D features} \label{table:thumos14_sota_i3d}
	    \vspace{-6pt}
    	\begin{tabular}{l | c c c c c}
    		\hline
    		{Method} & {0.3} & {0.4} & {0.5} & {0.6} & {0.7}\\
    		\hline 

            {W-TALC~\cite{paul_eccv_2018}} & 40.1 & {31.1} & 22.8 & {-} & 7.6 \\
    		{CMCS~\cite{liu2019completeness}} & {41.2} & {32.1} & {23.1} & {15.0} & {7.0} \\ 
            {TSM~\cite{yu2019temporal}} & {39.5} & {-} & {24.5} & {-} & {7.1} \\ 
            {3C-Net \cite{paul_eccv_2018}} & {40.9} & {32.3} & {24.6} & {-} & {7.7} \\
            {3C-Net$\dagger$ \cite{paul_eccv_2018}} & {44.2} & {34.1} & {26.6} & {-} &{8.1} \\
            {Nguyen \etal~\cite{nguyen2019weakly}} & \textbf{46.6} & \textbf{37.5} & {26.8} & {17.6} & {9.0} \\
            {BaS-Net~\cite{basnet_aaai20}} & {44.6} & {36.0} & \textbf{27.0} & \textbf{18.6} & \textbf{10.4} \\
            \hline
    		{\textbf{RefineLoc} ($\eta = 0$)} & {34.8} & {27.7} & {19.5} & {10.7} & {4.60} \\
            {\textbf{RefineLoc} ($\eta = 14$)} & {40.8} & {32.7} & {23.1} & {13.3} & {5.3} \\        
    		\hline
        \end{tabular}
    \end{subtable}
    \vspace{-6pt}
	\caption{\textbf{State-of-the-art Weak Supervision on THUMOS14}. RefineLoc is competitive using both feature types (tables \textbf{a} and \textbf{b}). $\dagger$ uses extra supervision from the number of instances per video.}
	\vspace{-12pt}
	\label{table:thumos14_sota}
\end{table}

\subsection{State-of-the-Art Comparison and Generalizability}\label{subsection:sota}

\noindent\textbf{On ActivityNet v1.2~\cite{dataset_activitynet} (Table~\ref{table:activitynet_sota})}. 
RefineLoc with TSN features outperforms state-of-the-art, CleanNet~\cite{liu2019weakly}, by $1.6\%$ in average mAP (Table~\ref{table:activitynet_sota_tsn}), while RefineLoc with I3D features shows competitive performance to BaS-Net~\cite{basnet_aaai20} (Table~\ref{table:activitynet_sota_i3d}). ActivityNet is large-scale and contains more diverse classes compared to THUMOS14. Thus, RefineLoc's strong performance on ActivityNet shows the effectiveness of our iterative refinement approach. We observe that our refinement process significantly enhances our base model, \ie RefineLoc ($\eta = 0$), by $9.9\%$ (TSN) and $13.5\%$ (I3D) in average mAP.

\noindent\textbf{On THUMOS14~\cite{dataset_thumos14} (Table~\ref{table:thumos14_sota})}. 
RefineLoc with TSN features (Table~\ref{table:thumos14_sota_tsn}) and with I3D features (Table~\ref{table:thumos14_sota_i3d}) exhibits competitive performance to state-of-the-art methods~\cite{basnet_aaai20,liu2019weakly,nguyen2019weakly}. We observe that our refinement process significantly enhances our base model, \ie RefineLoc ($\eta = 0$), by $19.7\%$ (TSN) and $3.6\%$ (I3D) in mAP@tIoU$=0.5$.

\noindent\textbf{On Generalizability (Table~\ref{table:generalization})}. We chose our WSTAL-base model to be simple, compared to other state-of-the-art models, to highlight the main contribution of our work, \ie the iterative refinement process. This process can lift the performance of such a simple model to compete and even outperform state-of-the-art methods on both datasets. Moreover, effectiveness of the refinement process is independent of the WSTAL-base model, which we demonstrate by generalizing our framework to other base models, namely W-TALC~\cite{paul_eccv_2018} and BaS-Net~\cite{basnet_aaai20}, on THUMOS14 using I3D features. These two methods employ attention-based models, where we apply our pseudo-background-foreground ground truth refinement process. Table~\ref{table:generalization} compares the results from the \textit{released code} of the two methods vs. their performance after adding our iterative refinement process. By doing this, we significantly improve both base methods. In fact, BaS-Net is improved by $1.77\%$ in mAP@tIoU$=0.5$, setting a new state-of-the-art performance on THUMOS14 ($28.03\%$ mAP@tIoU$=0.5$). Its important to note that the numbers obtained with the released codes differ from the ones reported in~\cite{paul_eccv_2018, basnet_aaai20}.

\begin{table}[t!]
    \small
	\centering
	\tabcolsep=0.07cm
	\begin{subtable}{\linewidth}
	    \tabcolsep=0.1cm
	    \centering
	    \vspace{3pt}
	    \caption{Generalizability of RefineLoc to other base models} 
	    \vspace{-6pt}
    	\begin{tabular}{l | c c c c c}
    		\hline
    		{Method} & {0.3} & {0.4} & {0.5} & {0.6} & {0.7}\\
    		\hline 

            {W-TALC Code~\cite{paul_eccv_2018}} & 42.98 & 34.59 & 26.99 & \textbf{17.74} & \textbf{9.42} \\
            {W-TALC Code + \textbf{RefineLoc}} & \textbf{44.10} & \textbf{35.08} & \textbf{27.66} & 17.67 & 9.14 \\
    		\hline 
        \end{tabular}
    \end{subtable}
    \begin{subtable}{\linewidth}
	    \tabcolsep=0.1cm
	    \centering
	    \vspace{3pt}
	    \caption{Generalizability of RefineLoc to other base models} 
	    \vspace{-6pt}
    	\begin{tabular}{l | c c c c c}
    		\hline
    		{Method} & {0.3} & {0.4} & {0.5} & {0.6} & {0.7}\\
    		\hline 
            {BaS-Net Code~\cite{basnet_aaai20}} & 43.40 & 35.16 & 26.26 & 18.59 & 10.16 \\
            {BaS-Net Code + \textbf{RefineLoc}}  & \textbf{45.10} & \textbf{36.50} & \textbf{28.03} & \textbf{18.95} & \textbf{10.36} \\
    		\hline
        \end{tabular}
    \end{subtable}
    \vspace{-6pt}
	\caption{\textbf{Generalizability of RefineLoc.} Our iterative refinement process generalizes to base models: (a) W-TALC~\cite{paul_eccv_2018} and (b) BasNet~\cite{basnet_aaai20}. We outperform W-TALC and BasNet baseline using 2 and 5 refinement iterations, respectively. By refining BasNet~\cite{basnet_aaai20}, we set a new state-of-the-art performance on THUMOS14.}
	\label{table:generalization}
	\vspace{-12pt}
\end{table}

We show that our method is simple, yet effective. We demonstrate that the key component of RefineLoc is the iterative process, showing its effectiveness regardless dataset, features, or base model. Despite its simplicity, RefineLoc outperforms all other methods using TSN features on ActivityNet, and beats the state-of-the-art when using BasNet and W-TALC as base models on THUMOS14. 

\begin{figure}[b]
  \centering
  \includegraphics[width=\linewidth]{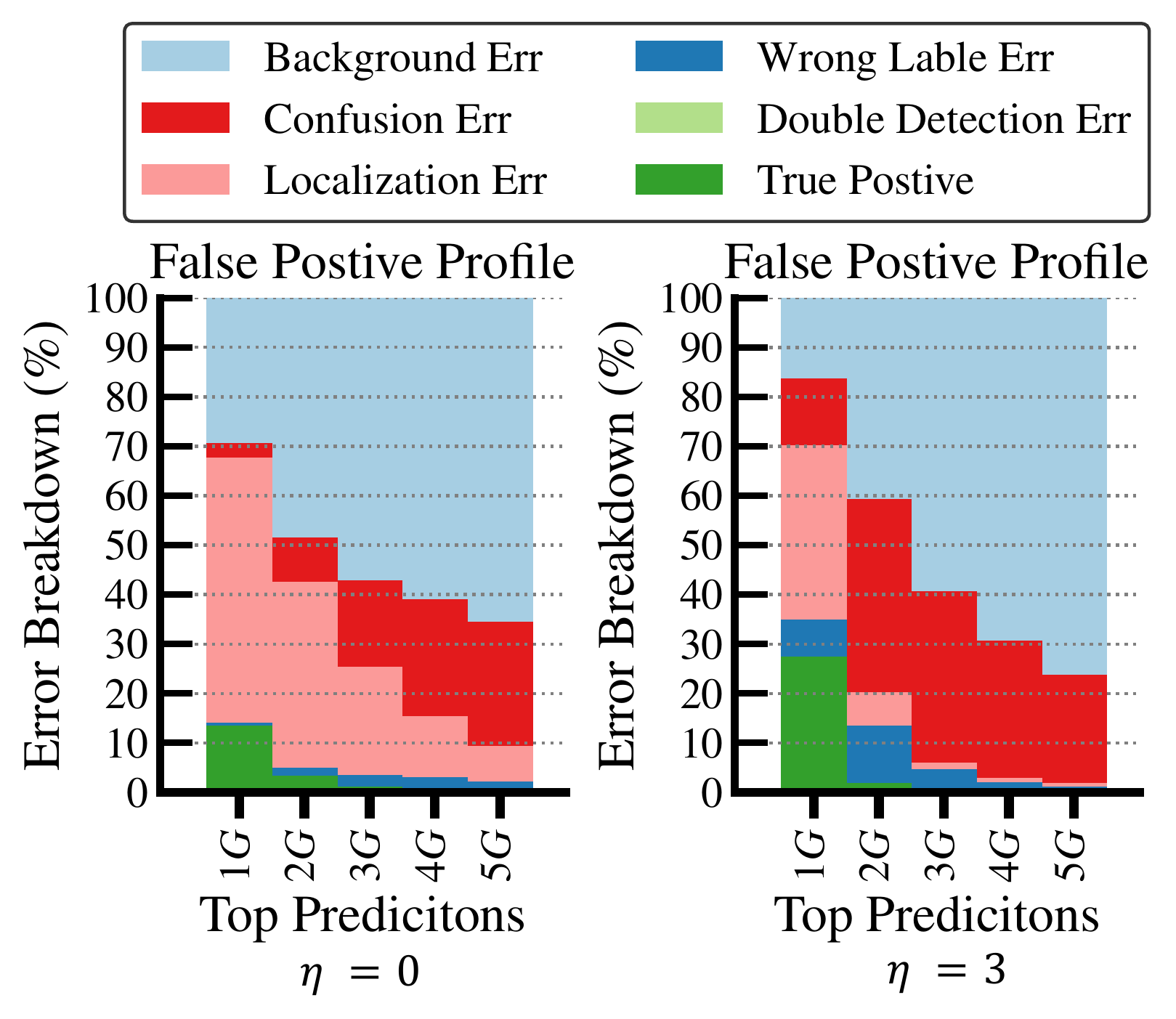}
  \caption{\textbf{Diagnosing Detection Results.} We present DETAD \cite{alwassel_eccv_2018} false positive profiles of RefineLoc at refinement iterations 0 and 3. $\textit{G}$ represents the number of ground truth segments available in the ActivityNet dataset. Our refinement strategy clearly pushes more true positive predictions to the top $1\textit{G}$ scoring predictions. RefineLoc also reduces background and localization error  at later iterations, indicating temporally tighter predictions.}
  \label{fig:detad_figure}
\end{figure}

\begin{figure*}[ht!]
  \centering
  \includegraphics[width=1\linewidth]{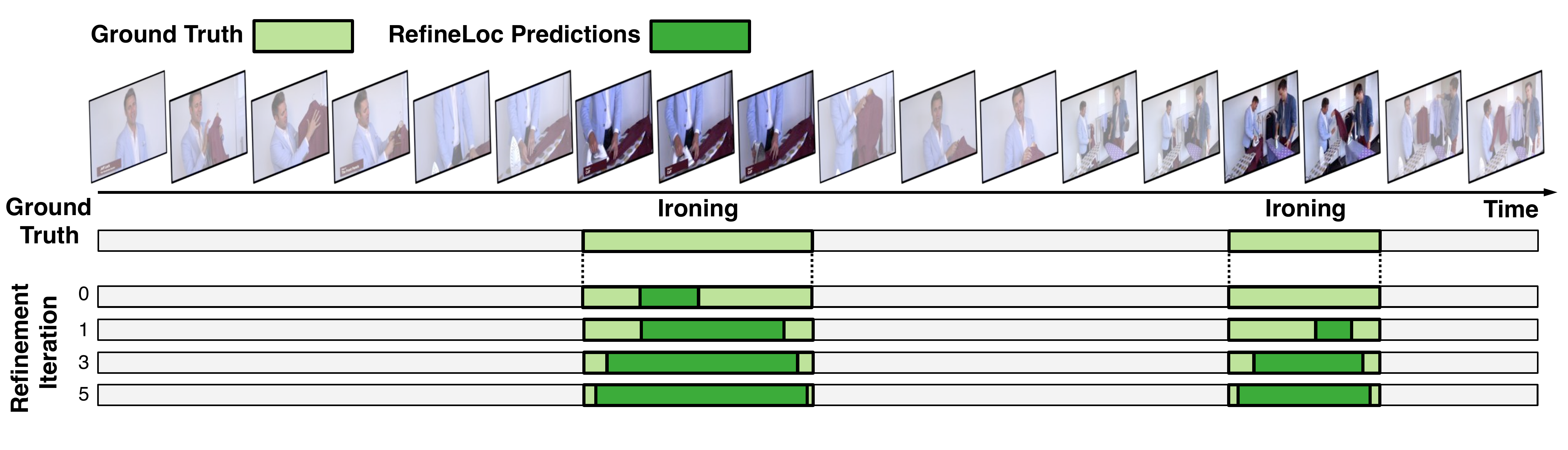}
  \includegraphics[width=1\linewidth]{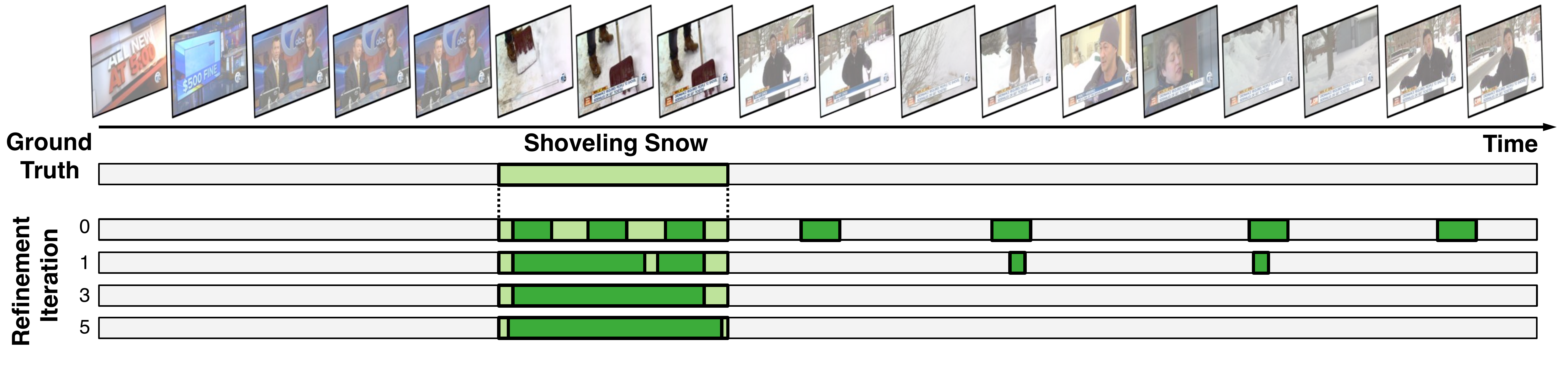}
  \includegraphics[width=1\linewidth]{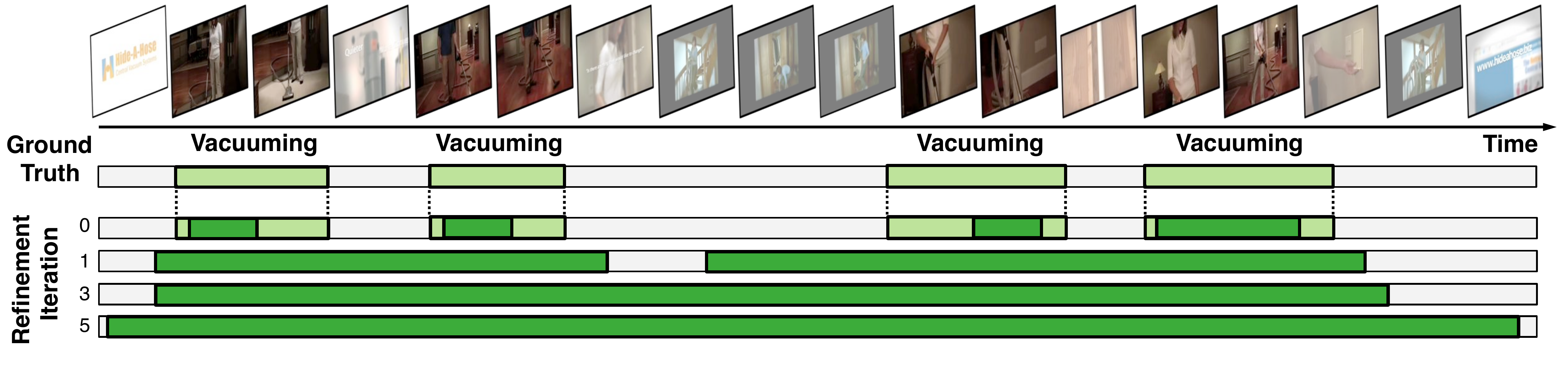}
  \caption{\textbf{Qualitative Results.} \textit{Top}: RefineLoc successfully enhances prediction coverage and detects missed instances as iterations evolve. \textit{Middle}: RefineLoc manages to merge disjoint predictions and remove wrong background predictions from one iteration to the next. \textit{Bottom}: In the presence of large context, iterative refinement can hurt RefineLoc predictions, as visual similarity between foreground and background confuses our attention model.}
  \label{fig:qualitative}
\end{figure*}

\subsection{Error Analysis and Qualitative Results}

\noindent\textbf{Diagnosing Detection Results.}
To analyze the merits of the proposed refinement strategy, we conduct a DETAD \cite{alwassel_eccv_2018} false-positive analysis of RefineLoc at refinement iterations 0 and 3. We present the results in Figure \ref{fig:detad_figure}. The false-positive profile analysis provides a fine-grained categorization of false-positive errors and summarizes the distribution of these errors over the top $5\textit{G}$ model predictions, where $\textit{G}$ is the number of ground truth segments in the dataset. After refinement (right plot), we observe that RefineLoc generates more high-scoring true positive predictions (towards $1\textit{G}$). Despite the reduction of background and localization errors, there is an increase in confusion errors. We explain this increase due to the simplicity of our initial classification module. Besides, the extra supervision generated by the pseudo-ground truth encourage the model to improve the localization but not directly the label prediction. 

\noindent\textbf{Qualitative Results}. Figure \ref{fig:qualitative} shows some RefineLoc qualitative detection results on ActivityNet. We present results for three different videos across different refinement iterations. The top video shows our method not only enhances its coverage over iterations, but is also able to detect a new instance at iteration $1$ that was missed in the previous iteration. In the middle video, we see how RefineLoc manages to successfully merge different predictions over iterations. We also see erroneous predictions being cut off from iteration to iteration. The final example shows a failure case. Despite starting with decent predictions at iteration $0$, our predictions diverge drastically in subsequent steps.

\section{Conclusion}
\label{section:conclusions}

We have presented RefineLoc, a novel weakly-supervised temporal action localization method. RefineLoc uses an iterative refinement strategy, where snippet-level pseudo labels are generated and used at every training iteration. Our experiments have shown that RefineLoc is competitive with the state-of-the-art and that our general iterative refinement process boosts the results of other methods outperforming the state-of-the-art, suggesting that it could be used as an \textit{off-the-shelf} strategy to refine results of future weakly-supervised methods for temporal action localization. As labeling videos for action localization is a massive time and cost bottleneck, RefineLoc takes a step closer to alleviating the need for these prohibitively expensive tasks.

\noindent\textbf{Acknowledgments.}
This work is supported the King Abdullah University of Science and Technology (KAUST) Office of Sponsored Research (OSR) under Award No. OSR-CRG2017-3405.

{\small
\bibliographystyle{ieee_fullname}
\bibliography{egbib}

\begin{thebibliography}{10}\itemsep=-1pt

\bibitem{alwassel_eccv_2018}
Humam Alwassel, Fabian Caba~Heilbron, Victor Escorcia, and Bernard Ghanem.
\newblock Diagnosing error in temporal action detectors.
\newblock In {\em ECCV}, 2018.

\bibitem{alwassel_2018_actionsearch}
Humam Alwassel, Fabian Caba~Heilbron, and Bernard Ghanem.
\newblock Action search: Spotting targets in videos and its application to
  temporal action localization.
\newblock In {\em ECCV}, 2018.

\bibitem{bilen_cvpr_2016}
Hakan Bilen and Andrea Vedaldi.
\newblock Weakly supervised deep detection networks.
\newblock In {\em CVPR}, 2016.

\bibitem{bojanowski_eccv_2014}
Piotr Bojanowski, R{\'e}mi Lajugie, Francis Bach, Ivan Laptev, Jean Ponce,
  Cordelia Schmid, and Josef Sivic.
\newblock Weakly supervised action labeling in videos under ordering
  constraints.
\newblock In {\em ECCV}, 2014.

\bibitem{buch_cvpr_2017}
Shyamal Buch, Victor Escorcia, Chuanqi Shen, Bernard Ghanem, and Juan
  Carlos~Niebles.
\newblock Sst: Single-stream temporal action proposals.
\newblock In {\em CVPR}, 2017.

\bibitem{caba_cvpr_2016}
Fabian Caba~Heilbron, Juan Carlos~Niebles, and Bernard Ghanem.
\newblock Fast temporal activity proposals for efficient detection of human
  actions in untrimmed videos.
\newblock In {\em CVPR}, 2016.

\bibitem{dataset_activitynet}
Fabian Caba~Heilbron, Victor Escorcia, Bernard Ghanem, and Juan~Carlos Niebles.
\newblock Activitynet: A large-scale video benchmark for human activity
  understanding.
\newblock In {\em CVPR}, 2015.

\bibitem{caron2018deepcluster}
Mathilde Caron, Piotr Bojanowski, Armand Joulin, and Matthijs Douze.
\newblock Deep clustering for unsupervised learning of visual features.
\newblock In {\em ECCV}, 2018.

\bibitem{caron2019self-pseudo}
Mathilde Caron, Piotr Bojanowski, Julien Mairal, and Armand Joulin.
\newblock Unsupervised pre-training of image features on non-curated data.
\newblock In {\em ICCV}, 2019.

\bibitem{i3d}
Joao Carreira and Andrew Zisserman.
\newblock Quo vadis, action recognition? a new model and the kinetics dataset.
\newblock In {\em CVPR}, 2017.

\bibitem{chang_cvpr_2019}
Chien-Yi Chang, De-An Huang, Yanan Sui, Li Fei-Fei, and Juan~Carlos Niebles.
\newblock D\({}^{\mbox{3}}\)tw: Discriminative differentiable dynamic time
  warping for weakly supervised action alignment and segmentation.
\newblock In {\em CVPR}, 2019.

\bibitem{chao_cvpr_2018}
Yu-Wei Chao, Sudheendra Vijayanarasimhan, Bryan Seybold, David~A Ross, Jia
  Deng, and Rahul Sukthankar.
\newblock Rethinking the faster r-cnn architecture for temporal action
  localization.
\newblock In {\em CVPR}, 2018.

\bibitem{cheron2018flexible}
Guilhem Ch{\'e}ron, Jean-Baptiste Alayrac, Ivan Laptev, and Cordelia Schmid.
\newblock A flexible model for training action localization with varying levels
  of supervision.
\newblock In {\em NeurIPS}, 2018.

\bibitem{dai_iccv_2017}
Xiyang Dai, Bharat Singh, Guyue Zhang, Larry~S Davis, and Yan Qiu~Chen.
\newblock Temporal context network for activity localization in videos.
\newblock In {\em ICCV}, 2017.

\bibitem{ding2018weakly}
Li Ding and Chenliang Xu.
\newblock Weakly-supervised action segmentation with iterative soft boundary
  assignment.
\newblock In {\em CVPR}, 2018.

\bibitem{duchenne_iccv_2009}
Olivier Duchenne, Ivan Laptev, Josef Sivic, Francis Bach, and Jean Ponce.
\newblock Automatic annotation of human actions in video.
\newblock In {\em ICCV}, 2009.

\bibitem{escorcia_arxiv_2018}
Victor Escorcia, Cuong~D Dao, Mihir Jain, Bernard Ghanem, and Cees Snoek.
\newblock Guess where? actor-supervision for spatiotemporal action
  localization.
\newblock {\em arXiv preprint arXiv:1804.01824}, 2018.

\bibitem{gaidon_ijcv_2013}
Adrien Gaidon, Zaid Harchaoui, and Cordelia Schmid.
\newblock Temporal localization of actions with actoms.
\newblock {\em IEEE transactions on pattern analysis and machine intelligence},
  2013.

\bibitem{action_segmentation}
J Gall and J Abu~Farha.
\newblock Ms-tcn: Multi-stage temporal convolutional network for action
  segmentation.
\newblock In {\em CVPR}, 2019.

\bibitem{gao_eccv_2018}
Jiyang Gao, Kan Chen, and Ram Nevatia.
\newblock Ctap: Complementary temporal action proposal generation.
\newblock In {\em ECCV}, 2018.

\bibitem{gao_iccv_2017}
Jiyang Gao, Zhenheng Yang, Kan Chen, Chen Sun, and Ram Nevatia.
\newblock Turn tap: Temporal unit regression network for temporal action
  proposals.
\newblock In {\em ICCV}, 2017.

\bibitem{activitynet_challenge}
Bernard Ghanem, Juan~Carlos Niebles, Cees Snoek, Fabian~Caba Heilbron, Humam
  Alwassel, Victor Escorcia, Ranjay Khrisna, Shyamal Buch, and Cuong~Duc Dao.
\newblock The activitynet large-scale activity recognition challenge 2018
  summary.
\newblock {\em arXiv preprint arXiv:1808.03766}, 2018.

\bibitem{huang_cvpr_2018}
De-An Huang, Shyamal Buch, Lucio Dery, Animesh Garg, Li Fei-Fei, and Juan
  Carlos~Niebles.
\newblock Finding "it": Weakly-supervised reference-aware visual grounding in
  instructional videos.
\newblock In {\em CVPR}, 2018.

\bibitem{huang_eccv_2016}
De-An Huang, Li Fei-Fei, and Juan~Carlos Niebles.
\newblock Connectionist temporal modeling for weakly supervised action
  labeling.
\newblock In {\em ECCV}, 2016.

\bibitem{dataset_thumos14}
Haroon Idrees, Amir~R Zamir, Yu-Gang Jiang, Alex Gorban, Ivan Laptev, Rahul
  Sukthankar, and Mubarak Shah.
\newblock The thumos challenge on action recognition for videos “in the
  wild”.
\newblock {\em Computer Vision and Image Understanding}, 2017.

\bibitem{Jain_2020_CVPR}
Mihir Jain, Amir Ghodrati, and Cees G.~M. Snoek.
\newblock Actionbytes: Learning from trimmed videos to localize actions.
\newblock In {\em CVPR}, 2020.

\bibitem{dataset_sports1m}
Andrej Karpathy, George Toderici, Sanketh Shetty, Thomas Leung, Rahul
  Sukthankar, and Li Fei-Fei.
\newblock Large-scale video classification with convolutional neural networks.
\newblock In {\em CVPR}, 2014.

\bibitem{dataset_kinetics}
Will Kay, Joao Carreira, Karen Simonyan, Brian Zhang, Chloe Hillier, Sudheendra
  Vijayanarasimhan, Fabio Viola, Tim Green, Trevor Back, Paul Natsev, et~al.
\newblock The kinetics human action video dataset.
\newblock {\em arXiv preprint arXiv:1705.06950}, 2017.

\bibitem{kuehne2017weakly}
Hilde Kuehne, Alexander Richard, and Juergen Gall.
\newblock Weakly supervised learning of actions from transcripts.
\newblock {\em Computer Vision and Image Understanding}, 2017.

\bibitem{kumar2017hide}
Krishna Kumar~Singh and Yong Jae~Lee.
\newblock Hide-and-seek: Forcing a network to be meticulous for
  weakly-supervised object and action localization.
\newblock In {\em ICCV}, 2017.

\bibitem{stips}
Ivan Laptev.
\newblock On space-time interest points.
\newblock {\em IJCV}, 2005.

\bibitem{laptev2008learning}
Ivan Laptev, Marcin Marszalek, Cordelia Schmid, and Benjamin Rozenfeld.
\newblock Learning realistic human actions from movies.
\newblock In {\em CVPR}, 2008.

\bibitem{basnet_aaai20}
Pilhyeon Lee, Youngjung Uh, and Hyeran Byun.
\newblock Background suppression network for weakly-supervised temporal action
  localization.
\newblock In {\em AAAI}, 2020.

\bibitem{lin_eccv_2018}
Tianwei Lin, Xu Zhao, Haisheng Su, Chongjing Wang, and Ming Yang.
\newblock Bsn: Boundary sensitive network for temporal action proposal
  generation.
\newblock In {\em ECCV}, 2018.

\bibitem{liu2019completeness}
Daochang Liu, Tingting Jiang, and Yizhou Wang.
\newblock Completeness modeling and context separation for weakly supervised
  temporal action localization.
\newblock In {\em CVPR}, 2019.

\bibitem{liu2019weakly}
Ziyi Liu, Le Wang, Qilin Zhang, Zhanning Gao, Zhenxing Niu, Nanning Zheng, and
  Gang Hua.
\newblock Weakly supervised temporal action localization through contrast based
  evaluation networks.
\newblock In {\em ICCV}, 2019.

\bibitem{ma2020sfnet}
Fan Ma, Linchao Zhu, Yi Yang, Shengxin Zha, Gourab Kundu, Matt Feiszli, and
  Zheng Shou.
\newblock {SF-Net: Single-Frame Supervision for Temporal Action Localization}.
\newblock In {\em ECCV}, 2020.

\bibitem{miech_iccv_2017}
Antoine Miech, Jean-Baptiste Alayrac, Piotr Bojanowski, Ivan Laptev, and Josef
  Sivic.
\newblock Learning from video and text via large-scale discriminative
  clustering.
\newblock In {\em ICCV}, 2017.

\bibitem{3cnet}
Sanath Narayan, Hisham Cholakkal, Fahad~Shahbaz Khan, and Ling Shao.
\newblock 3c-net: Category count and center loss for weakly-supervised action
  localization.
\newblock In {\em ICCV}, 2019.

\bibitem{neverova2019slim}
Natalia Neverova, James Thewlis, Riza~Alp Guler, Iasonas Kokkinos, and Andrea
  Vedaldi.
\newblock Slim densepose: Thrifty learning from sparse annotations and motion
  cues.
\newblock In {\em CVPR}, 2019.

\bibitem{nguyen_cvpr_2018}
Phuc Nguyen, Ting Liu, Gautam Prasad, and Bohyung Han.
\newblock Weakly supervised action localization by sparse temporal pooling
  network.
\newblock In {\em CVPR}, 2018.

\bibitem{nguyen2019weakly}
Phuc~Xuan Nguyen, Deva Ramanan, and Charless~C Fowlkes.
\newblock Weakly-supervised action localization with background modeling.
\newblock In {\em ICCV}, 2019.

\bibitem{oneata_cvpr_2014}
Dan Oneata, Jakob Verbeek, and Cordelia Schmid.
\newblock Efficient action localization with approximately normalized fisher
  vectors.
\newblock In {\em CVPR}, 2014.

\bibitem{oquab_cvpr_2015}
Maxime Oquab, L{\'e}on Bottou, Ivan Laptev, and Josef Sivic.
\newblock Is object localization for free?-weakly-supervised learning with
  convolutional neural networks.
\newblock In {\em CVPR}, 2015.

\bibitem{papandreou_iccv_2015}
George Papandreou, Liang-Chieh Chen, Kevin~P Murphy, and Alan~L Yuille.
\newblock Weakly-and semi-supervised learning of a deep convolutional network
  for semantic image segmentation.
\newblock In {\em ICCV}, 2015.

\bibitem{paul_eccv_2018}
Sujoy Paul, Sourya Roy, and Amit~K Roy-Chowdhury.
\newblock W-talc: Weakly-supervised temporal activity localization and
  classification.
\newblock In {\em ECCV}, 2018.

\bibitem{piergiovanni_cvpr_2018}
AJ Piergiovanni and Michael~S Ryoo.
\newblock Learning latent super-events to detect multiple activities in videos.
\newblock In {\em CVPR}, 2018.

\bibitem{richard_cvpr_2017}
Alexander Richard, Hilde Kuehne, and Juergen Gall.
\newblock Weakly supervised action learning with rnn based fine-to-coarse
  modeling.
\newblock In {\em CVPR}, 2017.

\bibitem{richard2018neuralnetwork}
Alexander Richard, Hilde Kuehne, Ahsan Iqbal, and Juergen Gall.
\newblock Neuralnetwork-viterbi: A framework for weakly supervised video
  learning.
\newblock In {\em CVPR}, 2018.

\bibitem{rolnick_arxiv_2017}
David Rolnick, Andreas Veit, Serge Belongie, and Nir Shavit.
\newblock Deep learning is robust to massive label noise.
\newblock {\em arXiv preprint arXiv:1705.10694}, 2017.

\bibitem{shi_iccv_2017}
Miaojing Shi, Holger Caesar, and Vittorio Ferrari.
\newblock Weakly supervised object localization using things and stuff
  transfer.
\newblock In {\em ICCV}, 2017.

\bibitem{shou_cvpr_2017}
Zheng Shou, Jonathan Chan, Alireza Zareian, Kazuyuki Miyazawa, and Shih-Fu
  Chang.
\newblock Cdc: Convolutional-de-convolutional networks for precise temporal
  action localization in untrimmed videos.
\newblock In {\em CVPR}, 2017.

\bibitem{shou_eccv_2018}
Zheng Shou, Hang Gao, Lei Zhang, Kazuyuki Miyazawa, and Shih-Fu Chang.
\newblock Autoloc: weakly-supervised temporal action localization in untrimmed
  videos.
\newblock In {\em ECCV}, 2018.

\bibitem{shou_cvpr_2016}
Zheng Shou, Dongang Wang, and Shih-Fu Chang.
\newblock Temporal action localization in untrimmed videos via multi-stage
  cnns.
\newblock In {\em CVPR}, 2016.

\bibitem{sigurdsson_iccv_2017}
Gunnar~A Sigurdsson, Olga Russakovsky, and Abhinav Gupta.
\newblock What actions are needed for understanding human actions in videos?
\newblock In {\em ICCV}, 2017.

\bibitem{two_stream}
Karen Simonyan and Andrew Zisserman.
\newblock Two-stream convolutional networks for action recognition in videos.
\newblock In {\em Advances in neural information processing systems}, 2014.

\bibitem{singh_iccv_2017}
Krishna~Kumar Singh and Yong~Jae Lee.
\newblock Hide-and-seek: Forcing a network to be meticulous for
  weakly-supervised object and action localization.
\newblock In {\em ICCV}, 2017.

\bibitem{dataset_ucf101}
Khurram Soomro, Amir~Roshan Zamir, and Mubarak Shah.
\newblock Ucf101: A dataset of 101 human actions classes from videos in the
  wild.
\newblock {\em arXiv preprint arXiv:1212.0402}, 2012.

\bibitem{tang_cvpr_2017}
Peng Tang, Xinggang Wang, Xiang Bai, and Wenyu Liu.
\newblock Multiple instance detection network with online instance classifier
  refinement.
\newblock In {\em CVPR}, 2017.

\bibitem{c3d}
Du Tran, Lubomir Bourdev, Rob Fergus, Lorenzo Torresani, and Manohar Paluri.
\newblock Learning spatiotemporal features with 3d convolutional networks.
\newblock In {\em ICCV}, 2015.

\bibitem{densetraj}
Heng Wang and Cordelia Schmid.
\newblock Action recognition with improved trajectories.
\newblock In {\em ICCV}, 2013.

\bibitem{wang_cvpr_2017}
Limin Wang, Yuanjun Xiong, Dahua Lin, and Luc Van~Gool.
\newblock Untrimmednets for weakly supervised action recognition and detection.
\newblock In {\em CVPR}, 2017.

\bibitem{tsn}
Limin Wang, Yuanjun Xiong, Zhe Wang, Yu Qiao, Dahua Lin, Xiaoou Tang, and Luc
  Van~Gool.
\newblock Temporal segment networks: Towards good practices for deep action
  recognition.
\newblock In {\em ECCV}, 2016.

\bibitem{xu_iccv_2017}
Huijuan Xu, Abir Das, and Kate Saenko.
\newblock R-c3d: Region convolutional 3d network for temporal activity
  detection.
\newblock In {\em ICCV}, 2017.

\bibitem{event_captioning}
Huijuan Xu, Kun He, Leonid Sigal, Stan Sclaroff, and Kate Saenko.
\newblock Text-to-clip video retrieval with early fusion and re-captioning.
\newblock In {\em AAAI}, 2019.

\bibitem{xu_cvpr_2015}
Jia Xu, Alexander~G Schwing, and Raquel Urtasun.
\newblock Learning to segment under various forms of weak supervision.
\newblock In {\em CVPR}, 2015.

\bibitem{xu2019gtad}
Mengmeng Xu, Chen Zhao, David~S. Rojas, Ali Thabet, and Bernard Ghanem.
\newblock G-tad: Sub-graph localization for temporal action detection.
\newblock In {\em CVPR}, 2020.

\bibitem{yeung_cvpr_2016}
Serena Yeung, Olga Russakovsky, Greg Mori, and Li Fei-Fei.
\newblock End-to-end learning of action detection from frame glimpses in
  videos.
\newblock In {\em CVPR}, 2016.

\bibitem{yu2019temporal}
Tan Yu, Zhou Ren, Yuncheng Li, Enxu Yan, Ning Xu, and Junsong Yuan.
\newblock Temporal structure mining for weakly supervised action detection.
\newblock In {\em ICCV}, 2019.

\bibitem{Zeng_2019_ICCV}
Runhao Zeng, Wenbing Huang, Mingkui Tan, Yu Rong, Peilin Zhao, Junzhou Huang,
  and Chuang Gan.
\newblock Graph convolutional networks for temporal action localization.
\newblock In {\em ICCV}, October 2019.

\bibitem{zhang_cvpr_2018}
Yongqiang Zhang, Yancheng Bai, Mingli Ding, Yongqiang Li, and Bernard Ghanem.
\newblock W2f: A weakly-supervised to fully-supervised framework for object
  detection.
\newblock In {\em CVPR}, 2018.

\bibitem{zhao_iccv_2017}
Yue Zhao, Yuanjun Xiong, Limin Wang, Zhirong Wu, Xiaoou Tang, and Dahua Lin.
\newblock Temporal action detection with structured segment networks.
\newblock In {\em ICCV}, 2017.

\end{thebibliography}
}

\appendix
\onecolumn

\section*{Supplementary Material}

\section{Additional Ablation Study}
Here, we include the same ablation study presented in the main paper (Subsection {\color{red} 4.3}) for three additional settings: ActivityNet v1.2~\cite{dataset_activitynet} using TSN~\cite{tsn} features and THUMOS14~\cite{dataset_thumos14} using TSN and I3D~\cite{i3d} features.

\noindent\textbf{Effects of the Pseudo Ground Truth Generator and the Loss Trade-off Coefficient $\beta$.}
Tables \ref{table:ablation_study_anet_tsn_generator_and_beta}, \ref{table:ablation_study_thumos14_tsn_generator_and_beta}, and \ref{table:ablation_study_thumos14_i3d_generator_and_beta} summarize the best performance for the five generators and for five different $\beta$ values on ActivityNet v1.2 using TSN, THUMOS14 using I3D, and THUMOS14 using TSN, respectively. The Segment Prediction-Based Generator consistently gives the best performance gain compared to the other generators in all settings. 

\begin{table}[h!]
    \small
	\centering
	\tabcolsep=0.15cm
    \begin{subtable}{1.0\linewidth}
        \centering
        \caption{ActivityNet v1.2 with TSN features}
    	\begin{tabular}{ l | c c c c c c}

    		\hline
    		
    		\textbf{Pseudo Ground}    & \multicolumn{6}{c}{\textbf{$\beta$}}\\
    		\textbf{Truth Generator}  & 0    & 1    & 2    & 4    & 8    & 16   \\
    		
    		\hline
                				
            Uniform Random & \multirow{5}{0.9em}{\rotatebox{90}{---\ \ 13.15\ \ ---}} & 13.15 & 13.15 & 13.15 &  13.15 & 13.15 \\
            Distribution Aware  &  & 15.27 & 19.22 & 18.76 & 20.80 & 20.96 \\
            Class Activation    &  & 22.95 & 22.90 & 22.53 & 22.55 & 22.23 \\
            Attention           &  & \textbf{23.15} &  22.90 & 22.47 & 22.57 & 22.36 \\
            Segment Prediction  &  & 23.09 & \underline{\textbf{23.16}} & \textbf{22.98} & \textbf{23.02} & \textbf{22.88} \\
    		\hline
    	\end{tabular}
    	\label{table:ablation_study_anet_tsn_generator_and_beta}
    \end{subtable}
    \begin{subtable}{\linewidth}
        \centering
        \caption{THUMOS14 with I3D features}
        \vspace{-5pt}
    	\label{table:ablation_study_thumos14_i3d_generator_and_beta}
    	\begin{tabular}{ l | c c c c c c}

    		\hline
    		
    		\textbf{Pseudo Ground}    & \multicolumn{6}{c}{\textbf{$\beta$}}\\
    		\textbf{Truth Generator}  & 0    & 1    & 2    & 4    & 8    & 16   \\
    		
    		\hline
                				
            Uniform Random & \multirow{5}{0.9em}{\rotatebox{90}{---\ \ 19.45\ \ ---}} & 21.12 & 20.20 & 19.78 & 19.45 & 19.45 \\
            Distribution Aware & & 20.69 & 20.32 & 19.45 & 19.45 & 19.45 \\
            Class Activation & & 20.18 & 20.11 & 20.10 & 20.21 & 20.34 \\
            Attention & & 19.45 & 19.45 & 19.45 & 19.45 & 19.45 \\
            Segment Prediction & & \textbf{21.48} & \underline{\textbf{22.60}} & \textbf{21.55} & \textbf{20.85} & \textbf{21.09} \\
    		\hline
    	\end{tabular}
    	
    \end{subtable}
    \begin{subtable}{\linewidth}
        \centering
    	\caption{THUMOS14 with TSN features}
        \vspace{-5pt}
		\label{table:ablation_study_thumos14_tsn_generator_and_beta}
    	\begin{tabular}{ l | c c c c c c}
    		\hline
    		
    		\textbf{Pseudo Ground}    & \multicolumn{6}{c}{\textbf{$\beta$}}\\
    		\textbf{Truth Generator}  & 0    & 1    & 2    & 4    & 8    & 16   \\
    		
    		\hline
                				
            Uniform Random & \multirow{5}{0.9em}{\rotatebox{90}{---\ \ 2.90\ \ ---}} & 17.97 & 17.64 & 18.60 & 18.96 & 16.64 \\
            Distribution Aware & & 14.89 & 14.73 & 14.90 & 16.42 & 14.50 \\
            Class Activation & & 12.12 & 12.32 & 13.66 & 12.98 & 13.28 \\
            Attention & & 20.70 & 21.37 & 21.10 & 20.64 & 19.66 \\
            Segment Prediction & & \textbf{20.92} & \textbf{21.87} & \underline{\textbf{22.63}} & \textbf{21.13} & \textbf{20.64} \\
    		\hline
    	\end{tabular}
    \end{subtable}
	\caption{\textbf{Effects of pseudo ground truth generator and loss trade-off coefficient $\beta$.} The metric is average mAP@tIoU=$0.5$:$0.05$:$0.95$ for ActivityNet v1.2 and mAP@tIoU$=0.5$ tIoU for THUMOS14. Bold represent the best generator for each $\beta$.}
\end{table}

\noindent\textbf{Performance over Refinement Iterations.}
Tables \ref{table:ablaiton_study_anet_tsn_iterations}, \ref{table:ablaiton_study_thumos14_i3d_iterations}, and \ref{table:ablaiton_study_thumos14_tsn_iterations} show the evolution of RefineLoc's performance across refinement iterations on ActivityNet v1.2 using TSN, THUMOS14 using I3D, and THUMOS14 using TSN. In each setting, we consistently observe a significant performance increase over our baseline model $\mathcal{M}_0$ (iteration $0$ in each table).

\begin{table}[t]
    \small
	\centering
	\tabcolsep=0.10cm
    \begin{subtable}{\linewidth}
        \centering
    	\caption{ActivityNet v1.2 using TSN features}
    	\label{table:ablaiton_study_anet_tsn_iterations}
    	\begin{tabular}{ l | c c c c c c}
    		\hline
    		
    		Refinement Iteration    & 0 & 1 & 2 & 3  & 4 & 5   \\
    		
    		\hline
                				
            \textbf{RefineLoc}  & 13.27  & 21.62 & 22.76 & 23.09 & 22.68 & \textbf{23.23} \\
    		\hline
    	\end{tabular}

    \end{subtable}
    \begin{subtable}{\linewidth}
        \centering
    	\caption{THUMOS14 using I3D features}
    	\label{table:ablaiton_study_thumos14_i3d_iterations}
    	\begin{tabular}{ l | c c c c c c}
    		\hline
    		
    		Refinement Iteration    & 0 & 3 & 6  & 9 & 12 & 14   \\
    		
    		\hline
                				
            \textbf{RefineLoc} & 19.45 & 20.96 & 21.36 & 22.46 & 21.87 & \textbf{23.12} \\
    		\hline
    	\end{tabular}
    \end{subtable}
    \begin{subtable}{\linewidth}
        \centering
    	\caption{THUMOS14 using TSN features}
    	\label{table:ablaiton_study_thumos14_tsn_iterations}
    	\begin{tabular}{ l | c c c c c c}
    		\hline
    		
    		Refinement Iteration    & 0    & 1    & 2    & 3    & 4    & 5   \\
    		
    		\hline
                				
            \textbf{RefineLoc} & 2.90 & 11.13 & 18.73 & 20.60 & \textbf{22.63} & 20.12 \\
    		\hline
    	\end{tabular}
    \end{subtable}
	\caption{\textbf{Performance over refinement iterations.} The reported metric is average mAP@tIoU=$0.5$:$0.05$:$0.95$ for ActivityNet v1.2 and mAP@tIoU$=0.5$ tIoU for THUMOS14. We observe that even with a weak base model, our method has the capability to improve the performance over iterations.}
\end{table}

\begin{figure}[t!]
    \begin{subfigure}{0.98\linewidth}
      \centering
      \includegraphics[trim={0 12.7cm 0 0},clip,width=\linewidth]{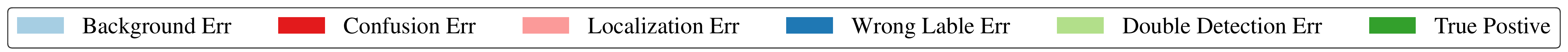}
    \end{subfigure}  
    \begin{subfigure}{0.48\linewidth}
      \centering
      \includegraphics[trim={0 0 0 3.2cm},clip,width=\linewidth]{figures/detad_anet_i3d.pdf}
      \vspace{-18pt}
      \caption{ActivityNet v1.2 using I3D features}
      \label{fig:detad_an_i3d_figure}
    \end{subfigure}
    \begin{subfigure}{0.48\linewidth}
      \centering
      \includegraphics[trim={0 0 0 3.2cm},clip,width=\linewidth]{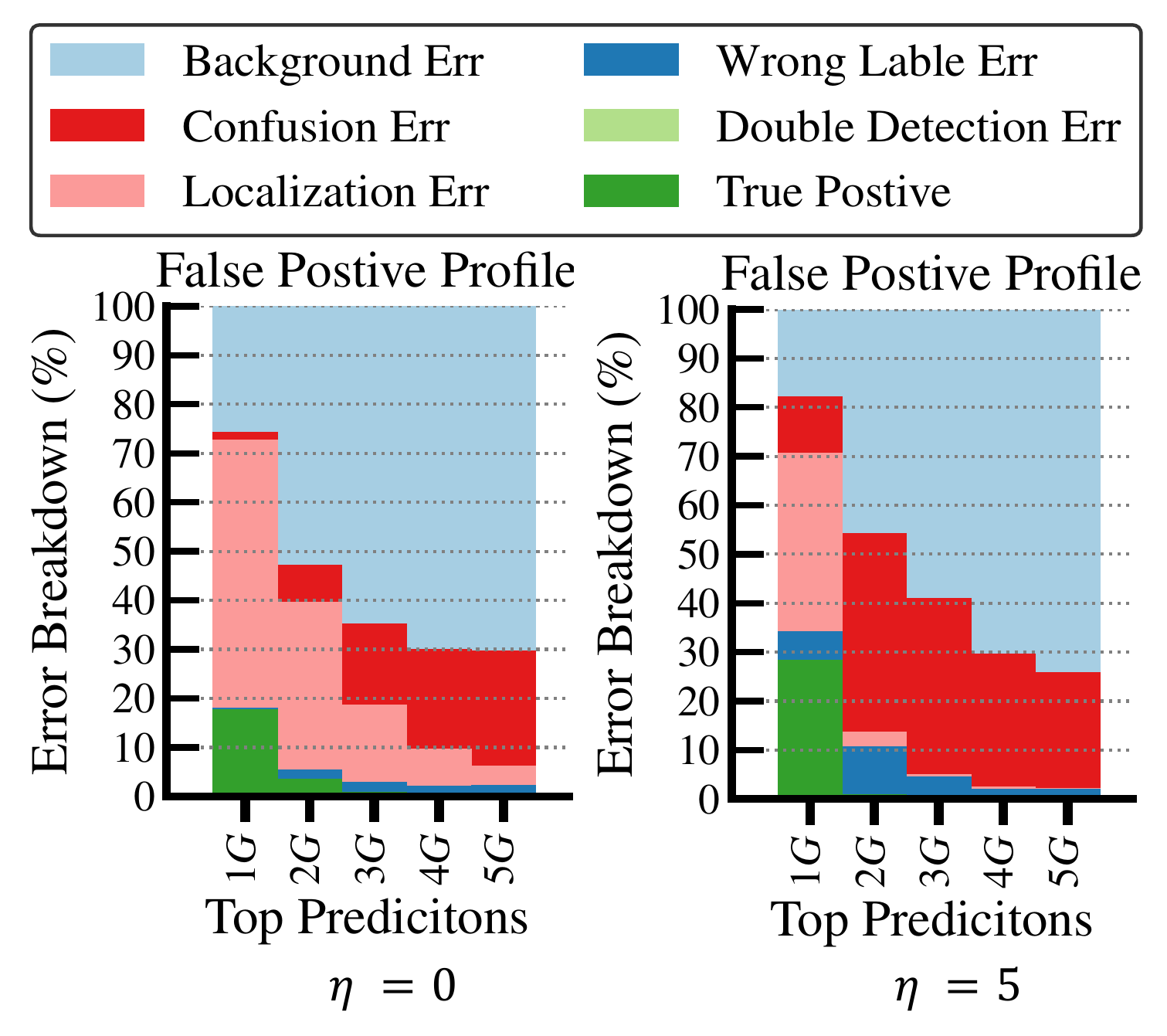}
      \caption{ActivityNet v1.2 using TSN features}
      \label{fig:detad_an_unt_figure}
    \end{subfigure}
    \begin{subfigure}{0.48\linewidth}
      \centering
      \includegraphics[trim={0 0 0 3.2cm},clip,width=\linewidth]{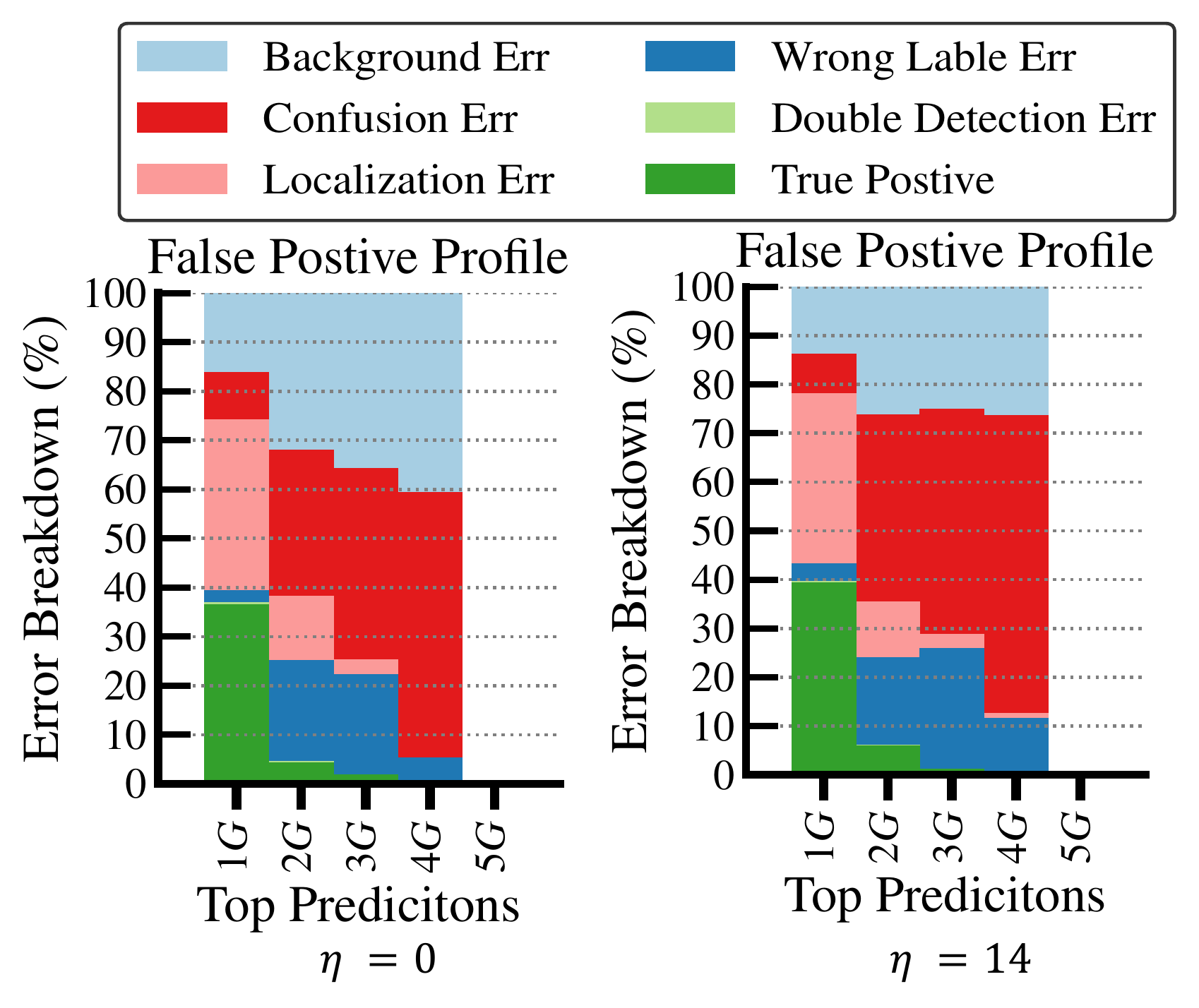}
      \caption{THUMOS14 using I3D features}
      \label{fig:detad_th_i3d_figure}
    \end{subfigure}
    \begin{subfigure}{0.48\linewidth}
      \centering
      \vspace{12pt}
      \includegraphics[trim={0 0 0 2.9cm},clip,width=\linewidth]{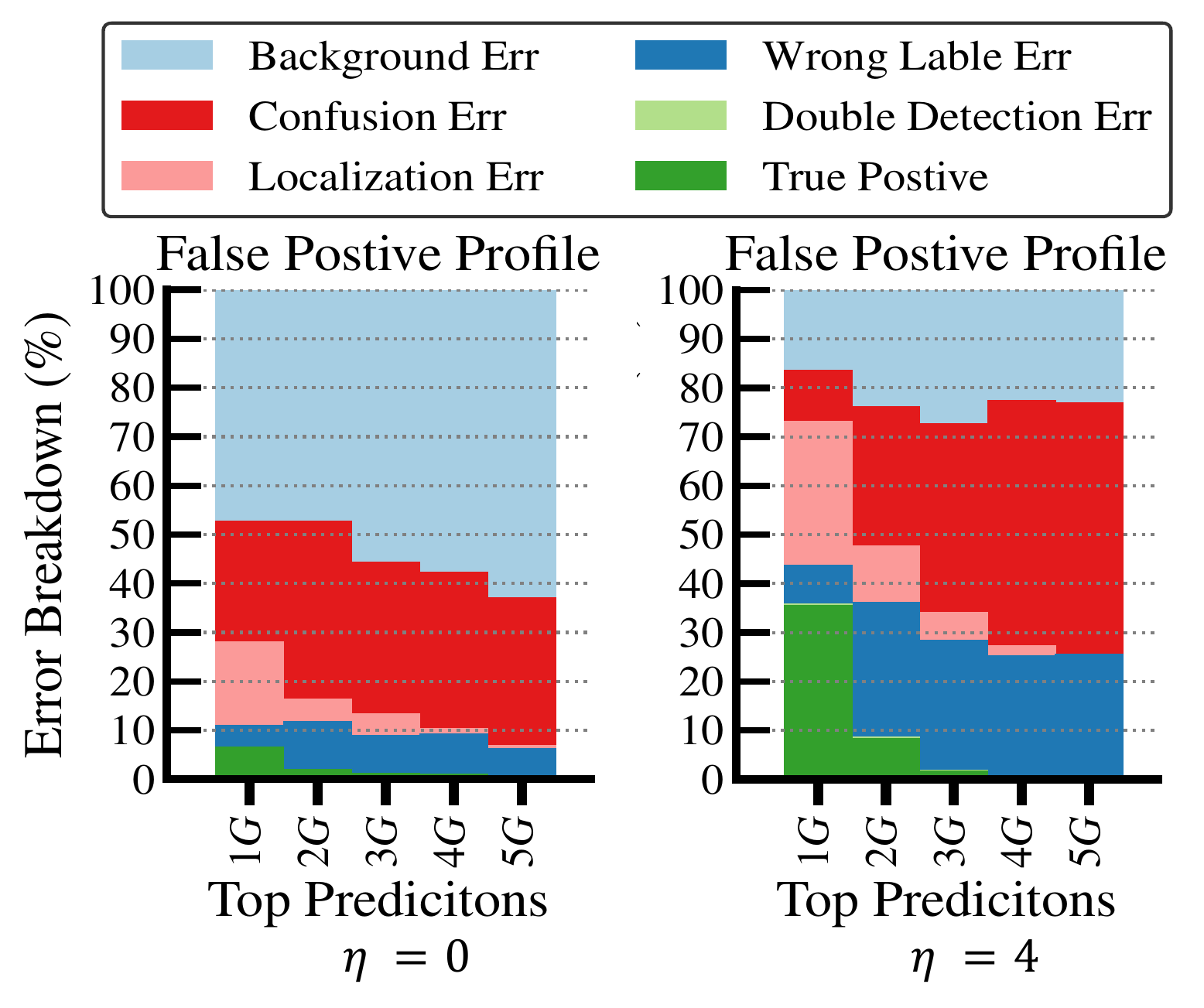}
      \vspace{-18pt}
      \caption{THUMOS14 using TSN features}
      \label{fig:detad_th_unt_figure}
    \end{subfigure}
    \vspace{12pt}
    \caption{\textbf{Diagnosing Detection Results.} We present DETAD \cite{alwassel_eccv_2018} false positive profiles of RefineLoc at refinement iterations $\eta=0$ (left) and $\eta=$ convergence (right). $\textit{G}$ represents the number of ground truth segments available in the dataset. Please refer to the DETAD paper~\cite{alwassel_eccv_2018} for the complete definition of each error type in the false positive profile.
    }
\end{figure}

\noindent\textbf{Diagnosing Detection Results.}
To further analyze the merits of the proposed refinement strategy, we conduct a DETAD \cite{alwassel_eccv_2018} false positive analysis of RefineLoc on ActivityNet v1.2 and THUMOS14 using I3D and TSN (Figures \ref{fig:detad_an_i3d_figure}, \ref{fig:detad_an_unt_figure}, \ref{fig:detad_th_unt_figure}, and \ref{fig:detad_th_i3d_figure}). The false-positive profile analysis provides a fine-grained categorization of false-positive errors and summarizes the distribution of these errors over the top $5\textit{G}$ model predictions, where $\textit{G}$ is the number of ground truth segments in the dataset. After refinement (right plot in each figure), we observe that RefineLoc generates more high-scoring true positive predictions (towards $1\textit{G}$) and reduces background and localization errors. The DETAD results indicate that our iterative refinement encourages tighter temporal predictions, which we argue does occur primarily because of the snippet-level supervision injected in the form of pseudo ground truth.

\section{Logistic Regression vs Cross-Entropy}

RefineLoc learns two values for the attention, instead of learning one single scalar. The motivation behind this design choice is to learn explicitly one value for background attention and one value for foreground attention. Besides, learning these two values trough a classification loss ($\ie$ cross-entropy) is an easier problem than learning one value through a regression loss ($\ie$ logistic regression). For ActivityNet, we found that our initial hypothesis is true. Indeed, when we learn only one scalar for attention, RefineLoc obtains only $22.2\%$ average mAP using I3D features, a $1\%$ drop in average mAP compared to the results obtained with cross-entropy. In contrast, the best result on THUMOS14 is obtained by learning only one scalar value. When learning two values for attention with cross-entropy, our model obtains only $19.95\%$ mAP at tIoU $0.5$.

\begin{figure}[ht!]
   \centering
   \includegraphics[width=1.0\linewidth]{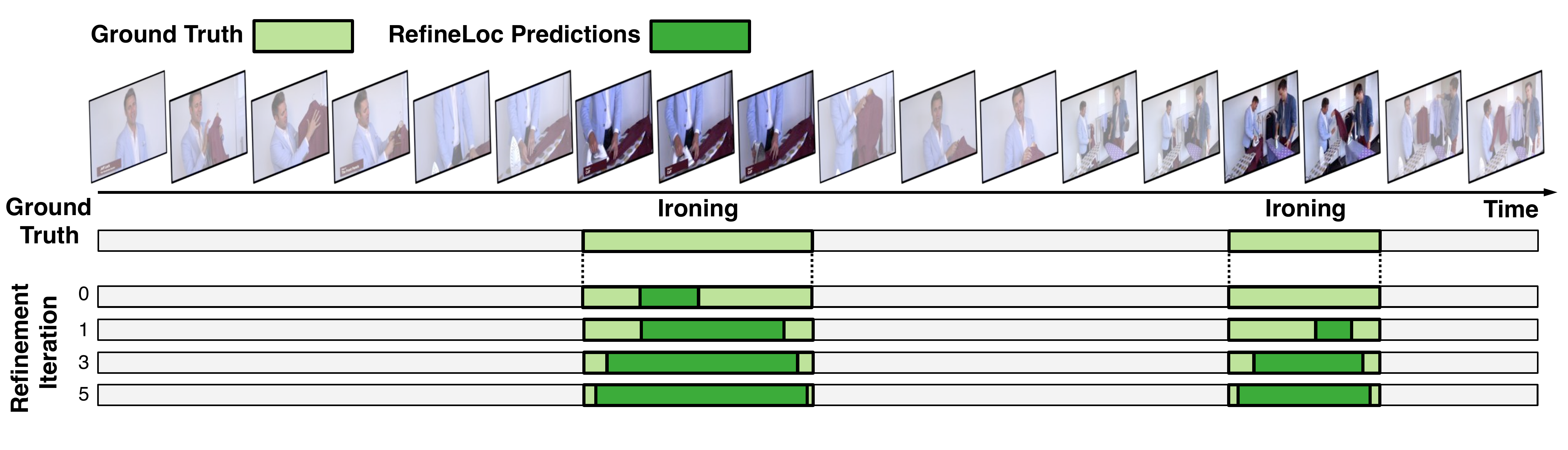}
   \includegraphics[width=1.0\linewidth]{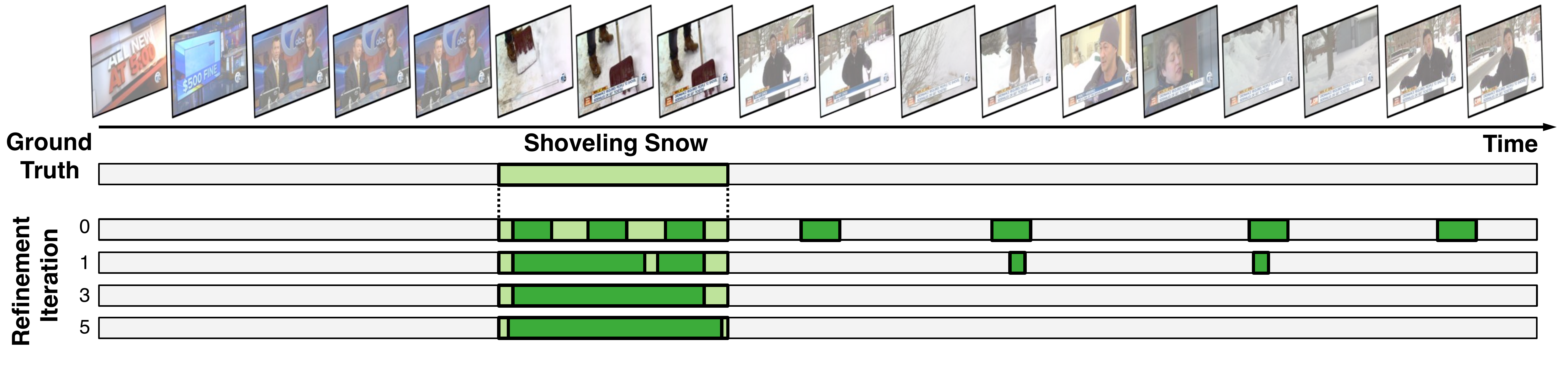}
   \includegraphics[width=1.0\linewidth]{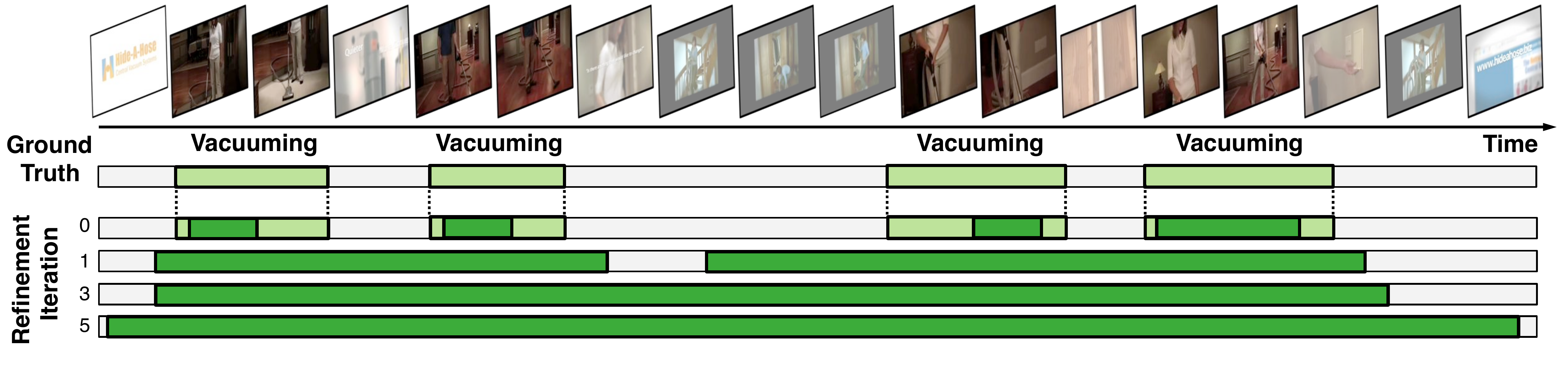}
   \caption{\textbf{Qualitative Results (ActivityNet v1.2).} \textit{Top}: RefineLoc successfully enhances prediction coverage and detects missed instances as iterations evolve. \textit{Middle}: RefineLoc manages to merge disjoint predictions and remove wrong background predictions from one iteration to the next. \textit{Bottom}: In the presence of large context, iterative refinement can hurt RefineLoc predictions, as visual similarity between foreground and background confuses our attention model.}
   \label{fig:qualitative_anet}
 \end{figure}

\subsection{Qualitative Results}
\noindent\textbf{ActivityNet v1.2}.
Figure \ref{fig:qualitative_anet} shows some RefineLoc qualitative detection results on ActivityNet. We present results across different refinement iterations. The top video shows our method not only enhances its coverage over iterations, but it is also able to detect a new instance at iteration $1$ that was missed in the previous iteration. In the middle video, we see how RefineLoc manages to successfully merge different predictions over iterations. We also see erroneous predictions being cut off from iteration to iteration. The final example shows a failure case. Despite the starting point at iteration $0$, our predictions diverge in later steps. We believe this confusion comes from the heavy context around the actions.

\begin{figure}[th!]
  \centering
  \includegraphics[width=1.0\linewidth]{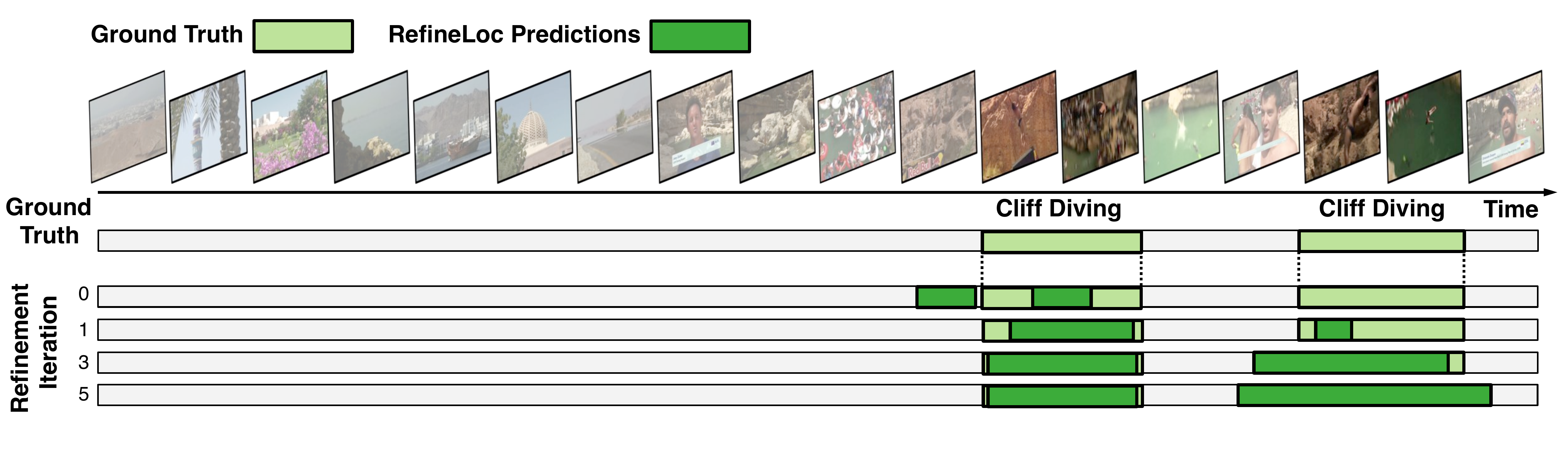}
  \includegraphics[width=1.0\linewidth]{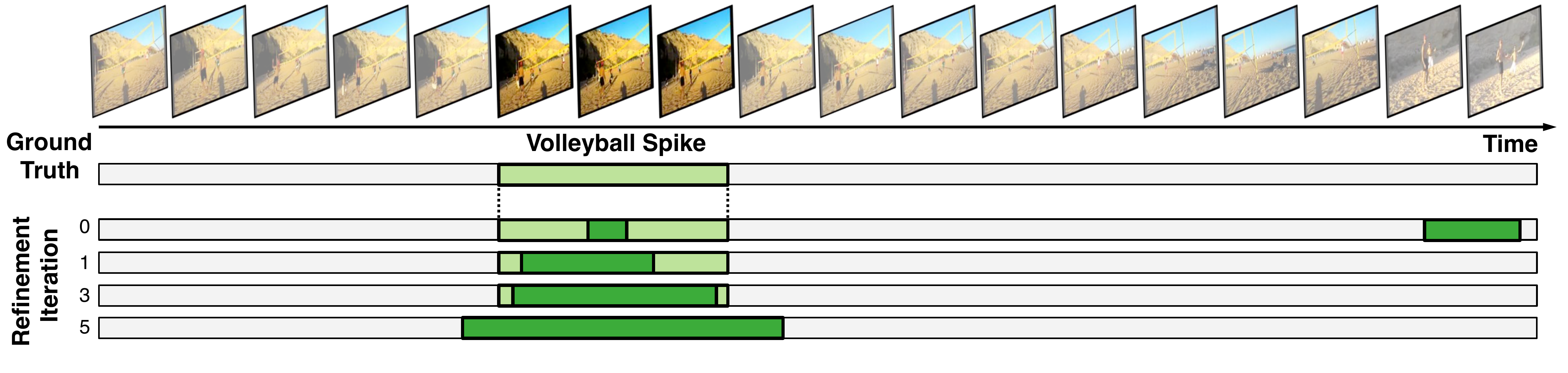}
  \includegraphics[width=1.0\linewidth]{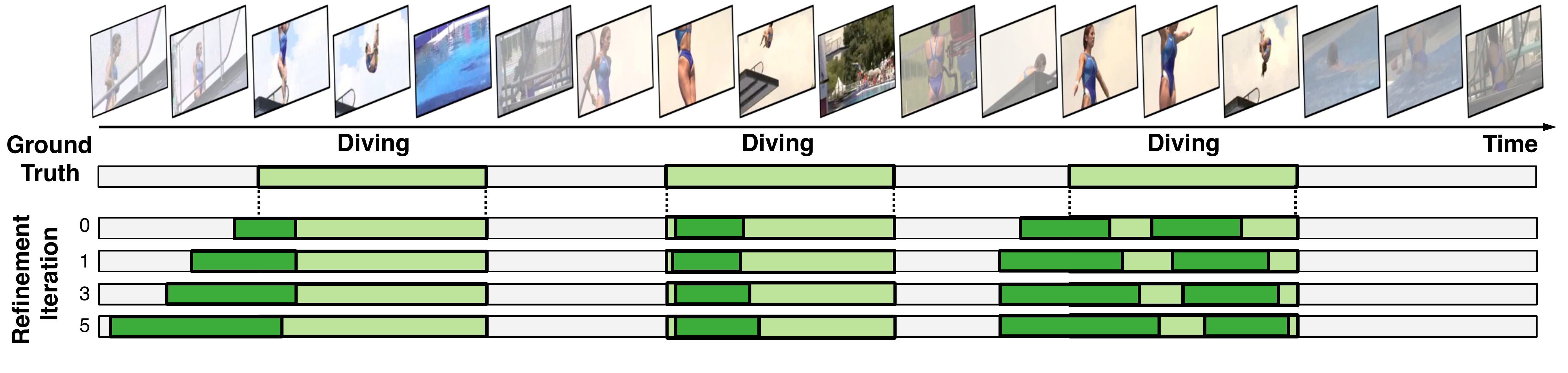}
  \caption{\textbf{Qualitative Results (THUMOS14).} \textit{Top and Middle}: RefineLoc successfully enhances prediction coverage over iterations and is able to detect missed instances as iterations evolve. \textit{Bottom}: In the presence of large context, iterative refinement can hurt RefineLoc predictions, as visual similarity between foreground and background confuses our attention model.}
  \label{fig:qualitative_thumos14}
\end{figure}

\noindent\textbf{THUMOS14}.
Figure \ref{fig:qualitative_thumos14} showcases RefineLoc qualitative results from the THUMOS14 dataset. We present results for three different videos over mulitple refinement iterations. The top video shows our method not only enhances its coverage over iterations, but it is also able to detect a new instance at iteration $1$ that was missed in the previous iteration. In the middle video, we see how RefineLoc manages to successfully cut off erroneous predictions from iteration to iteration. The final example shows a failure case. Despite starting with decent predictions at iteration $0$, our predictions do not improve in subsequent steps. We believe this confusion comes from the heavy context around the actions.

\end{document}